\documentclass[wcp,cleveref]{jmlr}

\usepackage{longtable}
\usepackage{booktabs}
\usepackage{multirow}
\usepackage{microtype}
\crefname{appendix}{Appendix}{Appendices}

\newcommand{\dnll}{\Delta\mathrm{NLL}}
\newcommand{\dr}{\Delta R^2}
\newcommand{\vcmax}{V_{c\max}}
\newcommand{\jmax}{J_{\max}}
\newcommand{\pn}{p/n}

\makeatletter
\let\Ginclude@graphics\@org@Ginclude@graphics
\makeatother

\jmlrvolume{353,}
\jmlryear{2026}
\jmlrworkshop{ACML 2026}

\title[Residual Correlation for GP Coregionalisation]{Residual Correlation as a Diagnostic for Joint-Uncertainty Gains from GP Coregionalisation}

\author[Zhou and Vanschoren]{\Name{Fangqin Zhou} \Email{f.zhou@tue.nl}\\
 \Name{Joaquin Vanschoren} \Email{j.vanschoren@tue.nl}\\
 \addr Eindhoven University of Technology, Eindhoven, The Netherlands}

\editors{Andy Song, Bo Han and Sarah Erfani}

\begin{document}

\makeatletter
\let \@jmlrpages \@empty
\makeatother

\maketitle

\begin{abstract}

In multi-target regression, correlated targets are often modelled using Gaussian processes with an intrinsic model of coregionalisation (GP-ICM), but target coupling does not consistently improve performance. Across the settings studied, we find that coupling primarily improves joint uncertainty rather than point prediction. Raw target correlation poorly predicts these gains; instead, residual correlation, the dependence remaining after fitting independent predictors, is more informative in the studied separable GP-ICM settings.

We introduce $D_{\text{logdet}} = -\tfrac{1}{2}\log\det\mathbf{R}_{\text{res}}$, a diagnostic that quantifies the idealised joint negative log-likelihood (NLL) gain from modelling a full rather than diagonal residual covariance, using independent GPs alone. Experiments on synthetic data, 16 multi-target benchmarks, and frozen transformer and convolutional neural network (CNN) representations for keypoint regression show little overall improvement in point prediction ($\Delta R^2 \approx 0$). Across the 16 benchmarks, $D_{\text{logdet}}$ strongly predicts observed ICM NLL improvements ($\rho_s = -0.83$, $p < 0.001$), outperforming raw target correlation and the feature-to-sample ratio. We also propose Residual-ICM, which incorporates residual correlation into the joint predictive covariance while preserving independent GP means and marginal variances. Residual-ICM achieves the best average joint NLL among the compared methods. Both $D_{\text{logdet}}$ and Residual-ICM target the global linear residual dependence captured by separable Gaussian covariance models.
\end{abstract}


\section{Introduction}
\label{sec:intro}
Multi-target regression is ubiquitous across the sciences \citep{Caruana1997,EvgeniouPontil2004,ZhangYang2022}. For instance, near-infrared spectra predict overlapping chemical constituents \citep{wold2001pls, pasquini2018near}, and shared environmental sensors estimate coupled water-quality indicators \citep{takada2023multi}. Similarly, in plant phenotyping, leaf spectra are used to jointly predict highly correlated capacities such as photosynthetic carboxylation rate ($\vcmax$) and electron transport ($\jmax$) \citep{serbin2012reflectance, DENG2024108540}. The standard machine learning pipeline for these tasks is to first use a fixed encoder to extract a representation $\Phi \in \mathbb{R}^{n \times p}$, followed by fitting a lightweight multi-output readout (prediction head) to the target matrix $\mathbf{Y} \in \mathbb{R}^{n \times T}$ \citep{ruder2017overview,radford2021learning}. However, deciding exactly how this readout should model the correlation between those targets remains a persistent structural challenge.

Correlated targets are commonly modelled jointly through coregionalised Gaussian processes (GPs), most often via the intrinsic model of coregionalisation (GP-ICM) \citep{Bonilla2007,AlvarezLawrence2011}, on the intuition that sharing information across related targets should improve prediction and uncertainty. Yet the benefit is inconsistent: some datasets gain substantially in joint uncertainty, whereas others gain little or even degrade despite strong target correlations \citep{rosenstein2005to, li2022negative}. The Tecator benchmark \citep{TecatorData}, for example, has a mean target correlation of $0.91$, yet coregionalisation increases joint negative log-likelihood (NLL) relative to independent GPs.

This raises a natural question: \emph{when does coupling help, and can this be predicted in advance?} Raw target correlation is the obvious candidate, but it does not determine whether a joint model has useful information to share. Coregionalisation acts on the uncertainty left after conditioning on the input representation. If independent per-target predictors already explain most target variation, little structure remains to exploit; if their prediction errors remain correlated, a joint model can improve uncertainty by modelling that residual dependence \citep{zellner1962efficient,Bonilla2007}.

We therefore focus on \emph{residual correlation}: the cross-target dependence remaining after fitting independent predictors. We formalise this dependence using $D_{\text{logdet}} = -\tfrac{1}{2}\log\det\mathbf{R}_{\text{res}}$, where $\mathbf{R}_{\text{res}}$ is the correlation matrix of standardised prediction residuals from independent GPs. The diagnostic can be computed without fitting a multi-output model and predicts the potential benefit of coregionalisation for joint uncertainty. However, because it relies on a single global Pearson correlation matrix, it captures only global linear residual dependence, consistent with the covariance structure modelled by separable GP-ICM. We examine its limitations under nonstationary and nonlinear dependence in Appendix~\ref{app:scope}.\footnote{All appendices are provided in the supplementary material.}


Across synthetic and real-world datasets, we examine how target coupling affects representation selection, point prediction, and joint uncertainty within a unified GP-readout framework. Our contributions are:

\begin{enumerate}

\item \textbf{Residual correlation as a diagnostic.}
We introduce $D_{\text{logdet}}$, computed from independent GPs, and show that it predicts joint-uncertainty gains from coregionalisation better than raw target correlation or $\pn$ ($\rho_s=-0.83$ versus $-0.07$ and $-0.46$).

\item \textbf{Empirical validation across settings.}
Using a synthetic phase diagram, 16 real datasets, and a sequential target-growth experiment, we show that coregionalisation gains concentrate in joint uncertainty estimation and increase only when correlated residual structure remains.

\item \textbf{Residual-ICM as a practical readout.}
We propose Residual-ICM, which incorporates residual correlation into the joint predictive covariance while preserving independent GP means and marginal variances. Across the 16 datasets, it achieves the best average joint NLL and mean rank among the compared methods.

\item \textbf{Consequences for selection and image encoders.}
Within the same framework, we show that target coupling adds little value for representation selection and that the residual-correlation mechanism also appears in frozen transformer and CNN representations for keypoint regression.
\end{enumerate}

\section{A Unified GP Readout Framework}
\label{sec:instrument}

\paragraph{Setup.}
Let $\Phi \in \mathbb{R}^{n \times p}$ denote a fixed feature matrix and $\mathbf{Y} \in \mathbb{R}^{n \times T}$ the target matrix, where $n$ is the number of samples, $p$ the feature dimension, and $T$ the number of targets. We compare independent and coregionalised GPs using the same features and held-out splits for representation selection, point prediction, and uncertainty quantification.

\paragraph{Single-output GP.}
For each target $t$, we fit an independent GP with a radial basis function (RBF) kernel $k(\mathbf{x}, \mathbf{x}') = \sigma_f^2 \exp(-\|\mathbf{x} - \mathbf{x}'\|^2 / (2\ell^2))$ (variance $\sigma_f^2$, lengthscale $\ell$) by maximum marginal likelihood \citep{RasmussenWilliams2006,MacKay1992}. The single-output GP log marginal likelihood (GP-LML) is our evidence for representation selection, and the resulting marginal predictions $p(y_t^* \mid \mathbf{x}^*, \Phi, \mathbf{y}_t)$ are used for marginal calibration.

\paragraph{Coregionalised GP (ICM).}
The Intrinsic Coregionalisation Model \citep{Bonilla2007,AlvarezLawrence2011} extends a scalar kernel $k_0$ to a multi-output joint covariance:
\begin{equation}
K_{\text{ICM}}\bigl((\mathbf{x},s),(\mathbf{x}',t)\bigr) = k_0(\mathbf{x},\mathbf{x}') \cdot B_{st}, \quad \mathbf{B} = \mathbf{W}\mathbf{W}^\top + \operatorname{diag}(\boldsymbol{\kappa}),
\label{eq:icm}
\end{equation}
where $\mathbf{W} \in \mathbb{R}^{T \times r}$ captures cross-target correlation and $\boldsymbol{\kappa}$ allows output-specific variation. 

\paragraph{Controlled comparison.}
To isolate the effect of target coupling, both independent GPs and ICM use the same input features. We fix the ICM lengthscale to the median optimised lengthscale of the independent GPs, use rank-1 coregionalisation, and cap the observation-noise variance at $\sigma^2 \le 0.5$. Ablations show that relaxing these constraints generally improves ICM, although some failures remain. We also evaluate higher ranks up to $T-1$ in \S\ref{sec:method}, with full ablations in Appendix~\ref{app:hardening}.

\paragraph{Baselines and evaluation.}
We compare against LogME \citep{YouLogME2021} for representation selection, and Bayesian linear regression (BLR) and deep ensembles \citep{LakshminarayananEnsembles2017} for predictive uncertainty. We use 5-fold cross-validation, or leave-one-out for datasets with $n<100$. For descriptive analyses, $D_{\text{logdet}}$ is computed from outer-fold out-of-fold predictions. When used for prediction or model selection, it is estimated using inner cross-validation within each training split, avoiding leakage from held-out data. \emph{Representation selection} is evaluated by Kendall $\tau$ between evidence-based and downstream $R^2$ rankings. \emph{Point prediction} is evaluated by $R^2$ and root-mean-square error (RMSE), with $\dr=R^2_{\text{ICM}}-R^2_{\mathrm{Indep}}$. \emph{Uncertainty} is evaluated by joint NLL, 90\%/95\% coverage gap, and sharpness, with $\dnll=\mathrm{NLL}_{\text{ICM}}-\mathrm{NLL}_{\text{Indep}}$; negative values favour ICM.

\paragraph{Joint NLL and predictive covariance.}
For each held-out sample $k$ with observed target vector $\mathbf{y}_k$, we compute the joint NLL under the full $T$-variate Gaussian predictive distribution $\mathcal{N}(\boldsymbol{\mu}_k,\boldsymbol{\Sigma}_k)$:
\begin{equation}
\mathrm{NLL}_k = \tfrac12\bigl[\,T\log 2\pi + \log\det\boldsymbol{\Sigma}_k + (\mathbf{y}_k-\boldsymbol{\mu}_k)^\top\boldsymbol{\Sigma}_k^{-1}(\mathbf{y}_k-\boldsymbol{\mu}_k)\,\bigr],
\label{eq:jointnll}
\end{equation}
We report the mean NLL across held-out samples, with targets standardised before fitting. Independent GPs use a diagonal predictive covariance, while ICM uses the full covariance. We add $10^{-8}\mathbf{I}$ jitter before Cholesky factorisation. To isolate the effect of cross-target dependence, $D_{\text{logdet}}$ uses a residual correlation matrix with unit diagonal, while Residual-ICM (\S\ref{sec:method}) preserves independent GP means and marginal variances, adding only cross-target covariance.

\section{The Residual-Correlation Diagnostic}
\label{sec:mechanism}

\paragraph{Residual correlation versus target correlation.}
Raw target correlation does not indicate how much cross-target dependence remains after conditioning on the input representation. Strongly correlated targets may be individually easy to predict, leaving little residual structure for coregionalisation to exploit. Conversely, moderately correlated targets may still have correlated prediction errors. Thus, we focus on \emph{residual correlation}, the cross-target dependence remaining after fitting independent predictors.

\paragraph{A residual-correlation diagnostic.}
Let $z_t(\mathbf{x}_i)=\bigl(y_{it}-\mu_t(\mathbf{x}_i)\bigr)/s_t(\mathbf{x}_i)$ denote the out-of-fold predictive-standardised residuals of independent GPs, where $\mu_t$ and $s_t$ are the predictive mean and standard deviation for target $t$. We define $\mathbf{R}_{\text{res}}=\operatorname{Corr}(z)$ as their $T\times T$ sample correlation matrix. Standardising by $s_t$ aligns the diagnostic with the residual-correlation structure used in Residual-ICM, whose predictive covariance is $\boldsymbol{\Sigma}=\operatorname{diag}(\mathbf{s})\mathbf{R}_{\text{res}}\operatorname{diag}(\mathbf{s})$ (\S\ref{sec:method}).

Under an idealised Gaussian setting with fixed marginal predictive variances, the expected reduction in joint negative log-likelihood (NLL) from modelling residual correlation rather than assuming independence is
\begin{equation}
D_{\text{logdet}} = -\tfrac{1}{2}\log\det \mathbf{R}_{\text{res}} \;\geq\; 0,
\label{eq:dlogdet}
\end{equation}
with equality if and only if $\mathbf{R}_{\text{res}}=\mathbf{I}$. The diagnostic is computed from independent GPs alone, with larger values indicating greater potential gains from covariance coupling. The realised ICM improvement may differ from this idealised gain because of finite-sample estimation, hyperparameter fitting, rank constraints, or non-Gaussian residuals.

\paragraph{Scope of the diagnostic.}
Because $D_{\text{logdet}}$ uses a single global Pearson correlation matrix, it cannot capture all forms of residual dependence. \Cref{tab:scope_tests} illustrates two limitations. First, input-dependent correlations that change sign across regions can cancel when pooled, leaving little benefit for global covariance coupling even when a region-wise model can exploit the dependence. Second, nonlinear but Pearson-uncorrelated residuals yield $D_{\text{logdet}}=0$ despite potential gains from nonparametric conditional modelling. Thus, a small $D_{\text{logdet}}$ indicates limited potential for global Gaussian covariance coupling, not statistical independence. Full experimental details are provided in Appendix~\ref{app:scope}.


\begin{table}[htbp]
\centering
\footnotesize
\caption{Scope stress tests of $D_{\text{logdet}}$ (three-seed means).
NLL differences are relative to the independent baseline
(negative favours coupling). The alternative is an oracle
region-wise covariance model for the input-dependent test
and a nonparametric conditional model for the nonlinear test.}
\label{tab:scope_tests}
\setlength{\tabcolsep}{6pt}
\begin{tabular}{llrrr}
\toprule
Test & Regime & $D_{\text{logdet}}$
& Global Gaussian $\Delta$NLL & Alternative $\Delta$NLL \\
\midrule
Input-dependent & Stationary    & $1.76$ & $-1.86$ & $-1.86$ \\
                & Nonstationary & $0.08$ & $-0.08$ & $-1.18$ \\
\midrule
Nonlinear       & Independent   & $0.00$ & $0.00$  & $+0.05$ \\
                & Linear        & $0.22$ & $-0.23$ & $-0.18$ \\
                & Nonlinear     & $0.00$ & $0.00$  & $-0.30$ \\
\bottomrule
\end{tabular}
\end{table}


\paragraph{Connection to observed ICM gains.}
We compare three predictors of the observed ICM benefit $\dnll$ across 16 datasets (\Cref{tab:datasets_full}). Raw target correlation $\bar\rho$ shows little association ($\rho_s=-0.07$, $p=0.80$), the feature-to-sample ratio $\pn$ is a moderate proxy ($\rho_s=-0.46$, $p=0.072$), and $D_{\text{logdet}}$ shows the strongest association ($\rho_s=-0.83$, $p<0.001$). Since $\dnll=\mathrm{NLL}_{\mathrm{ICM}}-\mathrm{NLL}_{\mathrm{Indep}}$, negative correlations indicate that larger diagnostic values are associated with greater ICM gains. 

Unlike the idealised gain represented by $D_{\text{logdet}}$, the observed $\dnll$ comes from a separately fitted rank-1 GP-ICM (\S\ref{sec:instrument}), so the two quantities need not coincide. Nevertheless, their association is strong (Pearson $\rho_p=-0.91$). It is also robust across datasets and seeds: the bootstrap 95\% confidence interval for Spearman $\rho_s$ is $[-0.99,-0.47]$, leave-one-dataset-out correlations range from $-0.92$ to $-0.80$, and per-seed correlations range from $-0.89$ to $-0.73$ (permutation $p<10^{-4}$). Similar behaviour appears for frozen transformer and CNN image representations (\S\ref{sec:images}); further robustness checks are reported in Appendix~\ref{app:robustness}.

\paragraph{Is the association explained by target count?}
Both $D_{\text{logdet}}$ and ICM gains tend to increase with the number of coupled targets, raising the possibility that their association is driven by target dimensionality. However, the association remains strong after normalising both quantities by $T$ ($\rho_s=-0.85$) and significant after controlling for $T$, $\pn$, and $R^2_{\mathrm{Indep}}$ (partial $\rho_s=-0.61$, $p=0.028$). Conversely, target count shows little remaining association with $\dnll$ after controlling for $D_{\text{logdet}}$ (partial $\rho_s=-0.05$, $p=0.85$; \Cref{tab:partial}). These results suggest that residual structure, not target count alone, explains the observed association.

\begin{table}[htbp]
\centering
\footnotesize
\caption{Separating residual structure from target dimensionality across the 16
datasets. Rank correlations are against $\dnll$, so negative values indicate
that a larger diagnostic is associated with a larger ICM improvement. Partial
correlations are rank-based, with $p$ computed at
$\mathrm{df}=n-2-k$ for $k$ controls.}
\label{tab:partial}
\setlength{\tabcolsep}{8pt}
\begin{tabular}{lcc}
\toprule
Statistic & Value & $p$ \\
\midrule
$\rho_s(D_{\text{logdet}},\dnll)$ (headline)
    & $-0.83$ & $<0.001$ \\
$\rho_s(\dnll/T,\;D_{\text{logdet}}/T)$ (per-target robustness)
    & $-0.85$ & $<0.001$ \\
partial $\rho_s$, controlling $T$
    & $-0.74$ & $0.002$ \\
partial $\rho_s$, controlling $T,\pn,R^2_{\mathrm{Indep}}$
    & $-0.61$ & $0.028$ \\
\midrule
partial $\rho_s(T,\dnll)$, controlling $D_{\text{logdet}}$
    & $-0.05$ & $0.85$ \\
\bottomrule
\end{tabular}
\end{table}

\paragraph{Why $\pn$ remains useful.}
The feature-to-sample ratio $\pn$ is a coarse, label-free proxy: when $p\gg n$, independent per-target models often leave larger and more correlated residuals, whereas when $p\ll n$, they often explain most target variation. However, $\pn$ does not directly measure residual dependence and is correspondingly a weaker predictor than $D_{\text{logdet}}$.

\paragraph{Ceiling-effect example.}
Tecator ($T=3$, $\pn=0.42$) illustrates why raw target correlation is insufficient. Despite strong target correlation (mean $0.91$), independent GPs already achieve high accuracy ($R^2_{\mathrm{Indep}}=0.945$), and ICM worsens joint NLL ($\dnll=+0.78$). Strong target dependence thus does not imply an available gain from coregionalisation.

\section{Empirical Validation}
\label{sec:validation}
We evaluate whether residual correlation predicts the benefit of coregionalisation, whether that benefit appears in uncertainty rather than point prediction, and how it changes as more targets are coupled. We first establish independent GPs as a calibrated baseline, then test the mechanism through a synthetic $(T,\pn)$ phase diagram, a 16-dataset real-data overlay, and a sequential target-growth experiment.

\subsection{Datasets}
\label{sec:datasets}

\paragraph{Hyperspectral plant physiology.}
We use three hyperspectral imaging datasets to predict the photosynthetic traits $\vcmax$ and $\jmax$ ($T=2$) from leaf spectra. SH and WUR are small-sample tomato datasets collected under multiple illumination conditions, whereas PHO \citep{DENG2024108540} is a larger mixed-canopy dataset. These datasets represent high-$\pn$ regimes where independent predictors may leave substantial residual uncertainty. 

\paragraph{Public multi-target benchmarks.}
We include public benchmarks from spectroscopy, environmental monitoring, hydrology, software-effort estimation, and robotics: tecator \citep{TecatorData}, energy, jura \citep{JuraData1993}, atp1d/atp7d, rf1/rf2, wq \citep{SpyromitrosWaterQuality2016}, scm1d/scm20d, edm, and sarcos. Together with SH, WUR, and the rice and wheat splits of PHO, these form 16 datasets covering $T=2$--$16$ and $\pn\approx0.03$--$5.0$ (\Cref{tab:datasets_full}). For computational comparability across the 16 datasets, larger datasets were subsampled to $n=280$ per seed; all Indep-versus-ICM comparisons use these subsamples.

\begin{table}[t]
\centering
\caption{Summary statistics for the 16 datasets. $\pn$: feature-to-sample ratio; $\bar{\rho}$: mean pairwise target correlation; $D_{\text{logdet}}$: residual-correlation diagnostic. $\dnll$ and $\dr$ denote ICM minus Indep, so negative $\dnll$ favours ICM. Values are averaged over five seeds, cross-validation split, and GP optimiser restart.}
\label{tab:datasets_full}
\setlength{\tabcolsep}{4pt}
\scriptsize
\begin{tabular}{lrrrrrrrrr}
\toprule
Dataset & $n$ & $p$ & $T$ & $\pn$ & $\bar\rho$ & $D_{\text{logdet}}$ & $\dnll$ & $\dr$ & $R^2_{\mathrm{Indep}}$ \\
\midrule
scm20d  & 280 &  61 & 16 & 0.22 & 0.58 & 6.83 & $-3.590$ & $-0.002$ & 0.546 \\
scm1d   & 280 & 280 & 16 & 1.00 & 0.65 & 3.92 & $-1.759$ & $-0.006$ & 0.791 \\
atp7d   & 280 & 411 &  6 & 1.47 & 0.63 & 2.02 & $-1.446$ & $-0.000$ & 0.530 \\
atp1d   & 280 & 411 &  6 & 1.47 & 0.82 & 1.57 & $-1.145$ & $-0.045$ & 0.715 \\
rf2     & 280 & 576 &  8 & 2.06 & 0.42 & 1.34 & $-0.851$ & $+0.031$ & 0.802 \\
rf1     & 280 &  64 &  8 & 0.23 & 0.40 & 1.25 & $-1.091$ & $+0.001$ & 0.927 \\
sarcos  & 280 &  21 &  7 & 0.07 & 0.37 & 1.01 & $+0.437$ & $-0.006$ & 0.933 \\
WUR     &  48 & 150 &  2 & 3.12 & 0.95 & 1.00 & $-0.800$ & $-0.175$ & 0.223 \\
SH      &  50 & 252 &  2 & 5.04 & 0.96 & 0.86 & $-0.535$ & $-0.143$ & 0.316 \\
rice    & 280 & 204 &  2 & 0.73 & 0.86 & 0.51 & $-0.133$ & $-0.008$ & 0.434 \\
tecator & 240 & 100 &  3 & 0.42 & 0.91 & 0.47 & $+0.778$ & $-0.008$ & 0.945 \\
wq      & 280 &  16 & 14 & 0.06 & 0.11 & 0.42 & $-0.140$ & $+0.007$ & 0.054 \\
wheat   & 280 & 204 &  2 & 0.73 & 0.88 & 0.35 & $-0.058$ & $+0.012$ & 0.504 \\
edm     & 154 &  16 &  2 & 0.10 & 0.01 & 0.03 & $+0.639$ & $+0.040$ & 0.438 \\
energy  & 280 &   8 &  2 & 0.03 & 0.82 & 0.03 & $-0.073$ & $+0.005$ & 0.839 \\
jura    & 280 &  15 &  3 & 0.05 & 0.20 & 0.01 & $-0.052$ & $-0.006$ & 0.638 \\
\midrule
\multicolumn{9}{l}{Spearman $\rho_s(\bar\rho,\; \dnll)$} & $-0.07$ \\
\multicolumn{9}{l}{Spearman $\rho_s(\log \pn,\; \dnll)$} & $-0.46$ \\
\multicolumn{9}{l}{Spearman $\rho_s(D_{\text{logdet}},\; \dnll)$} & $\mathbf{-0.83}$ \\
\bottomrule
\end{tabular}
\end{table}

\subsection{The Single-Output GP Is a Strong Calibrated Baseline}
\label{sec:calib}
Before evaluating target coupling, we verify that independent GPs provide a competitive baseline. Compared with Bayesian linear regression and deep ensembles, independent GPs achieve the lowest NLL and sharpest intervals on Tecator and Energy, and perform comparably to deep ensembles on PHO, with similar coverage (Appendix~\ref{app:calibration}). It also recovers residual cross-target structure better than linear and multilayer perceptron probes (Appendix~\ref{app:gateb}), supporting their use as the baseline for evaluating covariance coupling.


\subsection{Synthetic Phase Diagram and Real-Data Overlay}
\label{sec:prediction}

We generate synthetic data using three shared Gaussian latent factors, mapped to $p$-dimensional inputs through a nonlinear transformation. Targets are constructed from heterogeneous factor loadings with independent Gaussian noise. We fix $n=150$ and vary $T\in\{2,3,5,8\}$ and $p\in\{40,150,600\}$, averaging over five seeds. This design varies $\pn$ while keeping the intrinsic signal dimension fixed.

\begin{figure}[t]
\centering
\includegraphics[width=.96\textwidth]{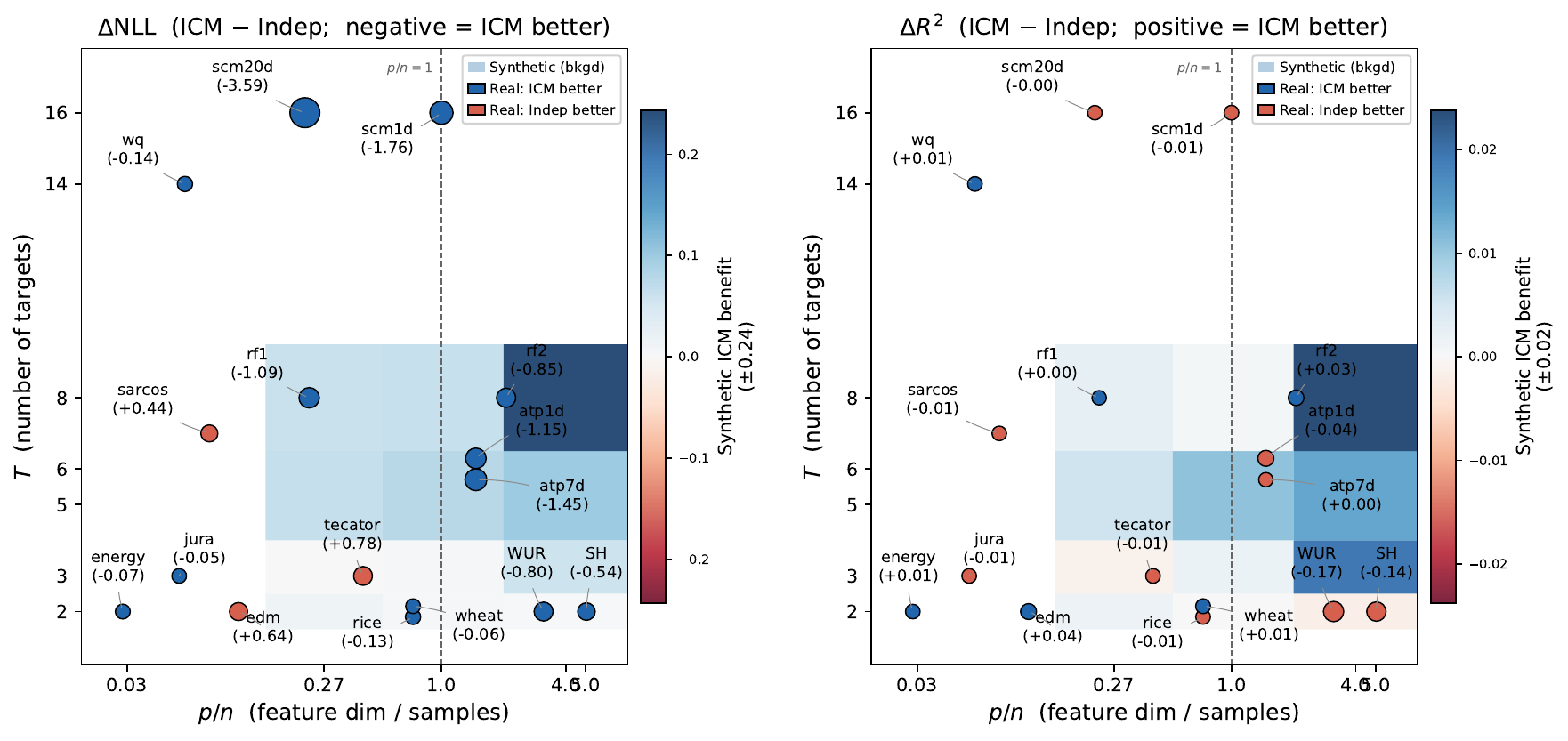}
\caption{Synthetic phase diagram with real-data overlay.
\textbf{Left:} Joint NLL difference $\dnll=\mathrm{NLL}_{\mathrm{ICM}}-\mathrm{NLL}_{\mathrm{Indep}}$ (negative favours ICM).
\textbf{Right:} Point-prediction difference $\dr=R^2_{\mathrm{ICM}}-R^2_{\mathrm{Indep}}$ (positive favours ICM). Background colours show the mean synthetic coupling benefit over $(T,\pn)$, with blue indicating larger gains ($-\dnll$ on the left and $\dr$ on the right).
Markers represent real datasets, coloured by the better-performing model, sized by effect magnitude, and labelled with signed differences. Colour scales are normalised separately for each panel.}
\label{fig:phase}
\end{figure}

\Cref{fig:phase} compares a controlled synthetic sweep over $(T,\pn)$ with results from 16 real datasets. In the synthetic experiments, joint NLL gains from coregionalisation generally increase with target count and $\pn$, while point-prediction changes remain near zero ($\dr\approx0$) across most settings. Varying target correlation preserves this pattern, with no coupling benefit when independent predictors leave no correlated residuals. Further experimental details and results are provided in Appendix~\ref{app:synthetic}.

The real datasets broadly follow the synthetic trend. Datasets such as scm20d, scm1d, atp1d, and atp7d show substantial joint NLL improvements. However, the variation across datasets confirms that $\pn$ is only a coarse proxy for coupling benefits. For example, Tecator shows worse joint NLL under ICM despite strong target correlation and high independent-GP accuracy. These results are consistent with the residual-correlation mechanism discussed in \S\ref{sec:mechanism}: coupling benefits depend on the residual dependence remaining after independent prediction.



\subsection{Sequential $T'$-Growth Experiment}
\label{sec:sequential}

\begin{figure}[htbp]
\centering
\includegraphics[width=.52\textwidth]{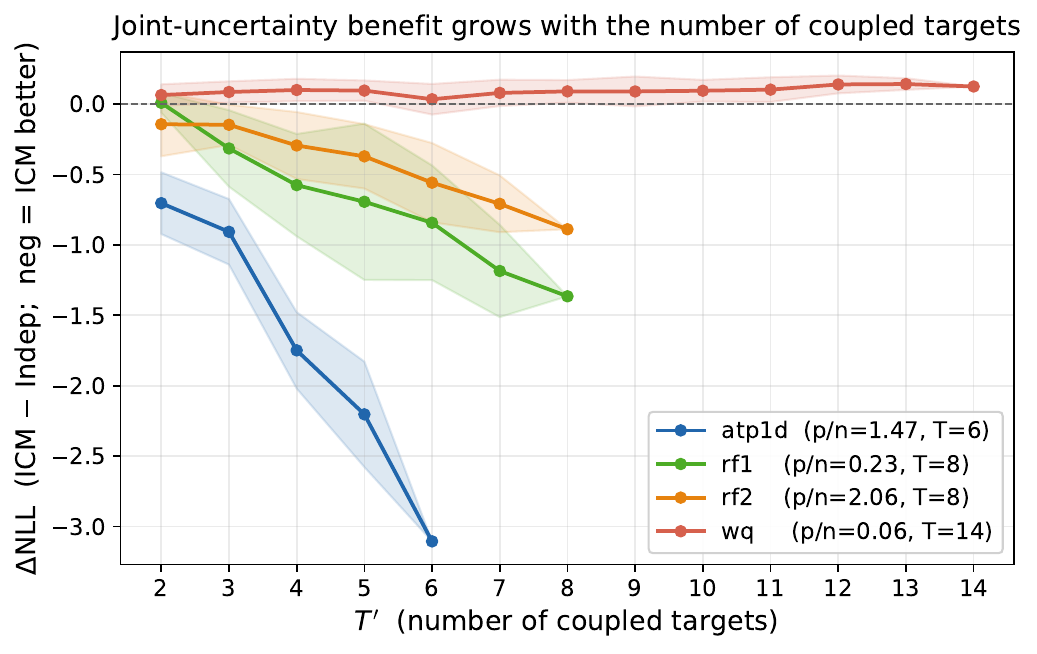}
\caption{Joint NLL difference $\dnll$ versus the number of coupled targets $T'$ (mean $\pm$ standard deviation over 20 random target subsets per $T'$). Negative values favour ICM. The benefit remains near zero for wq but increases with $T'$ for the others.}
\label{fig:sequential}
\end{figure}

\Cref{fig:sequential} shows three patterns. For atp1d, the ICM benefit increases nearly monotonically with $T'$. Similar trends appear for rf1 and rf2, although gains increase more slowly for rf2 despite its larger $\pn$, suggesting that target count and $\pn$ alone do not determine the benefit. In contrast, wq shows little improvement as targets are added, consistent with its limited residual dependence. Point-prediction changes remain small across all four datasets ($|\dr|\le0.05$). These results suggest that joint NLL gains accumulate with target count when exploitable residual dependence remains.

\section{From Diagnostic to Practical Algorithms}
\label{sec:method}

The diagnostic in \Cref{eq:dlogdet} suggests when target coupling may be useful. When residual dependence is weak, coupling may offer little benefit or even worsen joint uncertainty estimates. When substantial residual dependence remains, modelling it may improve joint NLL. We consider two strategies.

\paragraph{Residual-ICM.}
Rather than fitting a joint GP, Residual-ICM first fits independent single-output GPs and estimates the residual correlation matrix $\mathbf{R}_{\text{res}}$ from predictive-standardised out-of-fold residuals obtained by inner cross-validation on the training data. It then constructs the joint predictive covariance as
$\boldsymbol{\Sigma}=\operatorname{diag}(\mathbf{s})\mathbf{R}_{\text{res}}\operatorname{diag}(\mathbf{s})$,
where $\mathbf{s}$ contains the independent-GP predictive standard deviations. This preserves the independent GP means and marginal variances while adding cross-target covariance.

Residual-ICM uses the same residual-correlation construction as $D_{\text{logdet}}$ in \Cref{eq:dlogdet}. However, for prediction, the correlation matrix is estimated strictly within the training split to avoid leakage from held-out evaluation data.

The sample correlation matrix is positive semi-definite when residual variances are nonzero, yielding a valid predictive covariance. For numerical stability, we apply shrinkage toward the identity:
$\mathbf{R}_{\text{res}}\leftarrow(1-\lambda)\mathbf{R}_{\text{res}}+\lambda\mathbf{I}$,
where $\lambda=\min(1,T/m)$ and $m$ is the number of out-of-fold training residuals. In our tabular experiments, $m\gg T$, giving $\lambda\lesssim0.07$. When $m$ approaches or falls below $T$, this rule can shrink the correlation matrix nearly or entirely to the identity. We instead use Ledoit--Wolf shrinkage to stabilise $D_{\text{logdet}}$ in such regimes (Appendix~\ref{app:shrinkage}).

Given $m$ residual samples, estimating $\mathbf{R}_{\text{res}}$ costs $O(mT^2)$, followed by a one-time $O(T^3)$ factorisation and $O(T^2)$ joint scoring per test point. The main additional cost is obtaining inner-fold out-of-fold predictions. We do not claim an asymptotic advantage over ICM, which already uses a Kronecker eigendecomposition with $O(n^3+T^3)$ evidence computation (Appendix~\ref{app:hardening}). Instead, Residual-ICM avoids fitting a joint multi-output model and can be combined with other probabilistic predictors that provide predictive means and variances.

\paragraph{Diagnostic-gated coupling.}
As a simpler alternative, we fit rank-1 ICM only when $D_{\text{logdet}}>\tau$ and $R^2_{\mathrm{Indep}}<0.9$; otherwise, we retain independent GPs. Both criteria are evaluated on training data. The accuracy threshold is fixed in advance, while $\tau$ is selected by leave-one-dataset-out (LODO) cross-validation, minimising the mean $\dnll$ on the other 15 datasets. The selected thresholds are small ($\tau\le0.031$), so the accuracy ceiling usually determines whether coupling is applied. The gate achieves a mean $\dnll=-0.61$, similar to a fixed $\tau=0.5$ ($-0.64$). Threshold sensitivity is examined in Appendix~\ref{app:ceiling}.

\paragraph{Comparative results.}
\Cref{tab:method} shows that Residual-ICM achieves the best average $\dnll$ ($-1.13$) and mean rank ($1.94$) across the 16 datasets, outperforming rank-$(T-1)$ ICM ($-0.93$), the strongest ICM configuration in our sweep. Even selecting the best ICM rank for each dataset in hindsight gives a mean $\dnll=-0.94$. On the ten datasets with $T\geq3$, hindsight-rank ICM achieves $-1.40$, compared with $-1.54$ for Residual-ICM.

\begin{table}[t]
\centering
\scriptsize
\caption{Uncertainty gains ($\dnll$; lower is better) for independent GPs, rank-1 ICM, rank-$(T-1)$ ICM, Residual-ICM, and diagnostic-gated coupling, averaged over five seeds. Rank-$(T-1)$ is the highest-rank ICM configuration evaluated; the gate uses rank-1 ICM with the leave-one-dataset-out threshold.}
\label{tab:method}
\setlength{\tabcolsep}{7pt}
\begin{tabular}{lrrrrr}
\toprule
Dataset & Indep & ICM$_{r=1}$ & ICM$_{r=T-1}$ & Residual-ICM & Gate \\
\midrule
scm20d (strong)   & $\phantom{+}0.00$ & $-3.59$ & $-5.95$ & $-5.41$ & $-3.59$ \\
atp1d (strong)    & $\phantom{+}0.00$ & $-1.14$ & $-1.56$ & $-1.93$ & $-1.14$ \\
rf1 ($R^2_{\mathrm{Indep}}{=}0.93$) & $\phantom{+}0.00$ & $-1.09$ & $-1.21$ & $-0.86$ & $\phantom{+}0.00$ \\
sarcos (ceiling)  & $\phantom{+}0.00$ & $+0.44$ & $-0.06$ & $-0.83$ & $\phantom{+}0.00$ \\
tecator (ceiling) & $\phantom{+}0.00$ & $+0.78$ & $+0.62$ & $-0.45$ & $\phantom{+}0.00$ \\
wq (low struct.)  & $\phantom{+}0.00$ & $-0.14$ & $-0.06$ & $-0.10$ & $-0.14$ \\
\midrule
\textbf{Mean $\dnll$} (16 sets) & $\phantom{+}0.00$ & $-0.61$ & $-0.93$ & $\mathbf{-1.13}$ & $-0.61$ \\
\textbf{Mean rank} (1=best)     & $4.34$ & $2.97$ & $2.56$ & $\mathbf{1.94}$ & $3.19$ \\
\bottomrule
\end{tabular}
\end{table}

Residual-ICM also has lower downside risk. Across the $T\geq3$ datasets, hindsight-rank ICM accumulates $0.62$ nats of positive $\dnll$, compared with $0.01$ for Residual-ICM (Appendix~\ref{app:rankcomp}). On Tecator, ICM worsens joint NLL across the tested ranks; on Sarcos, increasing the rank largely resolves the rank-1 failure. Residual-ICM improves joint NLL on both datasets while preserving independent GP marginal predictions.

The diagnostic gate is more conservative. It avoids the rank-1 ICM failures on Tecator and Sarcos but also misses the improvement on rf1 ($R^2_{\mathrm{Indep}}=0.93$), where Residual-ICM achieves $\dnll=-0.86$. Although the gate does not improve mean NLL over always fitting rank-1 ICM (both $-0.61$), it reduces downside risk: rank-1 ICM worsens NLL on three datasets by a total of $1.85$ nats, whereas the gate couples on 11 of 16 datasets and worsens NLL on only one by $0.64$ nats. Thus, the gate serves as a risk-aware screening rule rather than a better performance selector. Residual-ICM remains the stronger choice for joint uncertainty in the settings studied.

To test whether the residual-covariance mechanism extends beyond GPs, we also evaluate a neural predictor on the same 16 datasets (Appendix~\ref{app:neural}). Adding post-hoc residual covariance improves joint NLL on 13 datasets (mean $\dnll=-0.80$), with gains strongly associated with the neural residual diagnostic ($\rho_s=-0.88$, $p<10^{-4}$). Jointly learning a full covariance gives a smaller average improvement ($\dnll=-0.34$) and reduces $R^2$ on 12 datasets. These results suggest that residual-covariance modelling can benefit other probabilistic predictors, although the evidence is limited to one neural architecture.

\paragraph{Decision Map.}
\Cref{tab:decision} summarises the practical implications. Within the separable GP-ICM settings studied, target coupling generally offers little benefit for representation selection or point prediction, but can improve joint uncertainty when residual dependence remains. The feature-to-sample ratio $\pn$ provides a coarse indicator before fitting, while $D_{\text{logdet}}$ and $R^2_{\mathrm{Indep}}$ can guide decisions after fitting independent GPs. High prediction accuracy does not necessarily rule out gains from Residual-ICM, as illustrated by rf1.

\begin{table}[htbp]
\centering
\scriptsize
\caption{Practical decision map for the separable GP-ICM setting studied here.
The $R^2_{\mathrm{Indep}}>0.9$ ceiling applies only to the diagnostic gate;
Residual-ICM can still benefit high-accuracy datasets when substantial residual
dependence remains. Recommendations reflect average behaviour rather than guarantees.}
\label{tab:decision}
\setlength{\tabcolsep}{5pt}
\begin{tabular}{llllp{7.6cm}}
\toprule
Goal & $T$ & $\pn$ & $R^2_{\mathrm{Indep}}$ & Recommendation \\
\midrule
Selection
& any & any & any
& Single-output GP-LML \\

Point prediction
& any & any & any
& Independent GPs \\

Joint NLL / calibration
& $\geq 2$ & $\geq 0.2$ & $<0.9$
& Residual-ICM; gains grow with $T$ and $\pn$ (modest for ${T\leq5}$), and a jointly fitted GP-ICM is competitive at large $T$ \\

Joint NLL / calibration
& any & $<0.2$ & any
& Independent GPs unless $D_{\text{logdet}}$ is large \\

Joint NLL / calibration
& any & any & $>0.9$
& Gate retains independent GPs; Residual-ICM may still help when
$D_{\text{logdet}}$ is large \\
\bottomrule
\end{tabular}
\end{table}

\section{Further Consequences: Selection and Image Encoders}
\label{sec:consequences}
We further examine whether target coupling improves representation selection and whether the residual-correlation mechanism extends to frozen transformer and CNN image representations.

\subsection{Representation Selection}
\label{sec:selection}

Within separable GP-ICM, multi-output evidence can be decomposed into single-output GP evidence after rotating the target space (proposition and proof in Appendix~\ref{app:proof}). 
When the coregionalisation matrix $\mathbf{B}$ is fixed across representations, the rotation and weighting of the eigen-target directions are also fixed. However, unequal eigenvalues can still change the representation rankings. Exact agreement with independent-GP evidence holds when $\mathbf{B}=\lambda\mathbf{I}$. In our experiments, $\mathbf{B}$ is re-optimised for each representation, so ranking invariance is not guaranteed. These results do not generally extend to non-separable multi-output GPs.


We evaluate representation selection on three datasets (PHO, Tecator, and Energy), with two prediction directions per dataset. For each of the six dataset--task pairs, we rank six representations: raw features, PCA-5, PCA-20, and MLP embeddings with widths of 32, 64, and 128. We measure ranking agreement using Kendall's $\tau$ between evidence-based rankings and held-out $R^2$ rankings. \Cref{tab:selection} shows that single-output GP marginal likelihood achieves the highest mean Kendall $\tau$ (0.706), outperforming GP-ICM (0.607) and LogME (0.527). Thus, target coupling provides little additional benefit for representation selection within the separable GP-ICM framework.

\begin{table}[htbp]
\centering
\scriptsize
\caption{Representation-selection performance measured by Kendall $\tau$ between evidence-based rankings and downstream $R^2$ rankings. Higher values indicate better agreement.}
\label{tab:selection}
\setlength{\tabcolsep}{10pt}
\begin{tabular}{lccc}
\toprule
Method & Mean $\tau$ & Best $\tau$ & Worst $\tau$ \\
\midrule
GP-LML (single-output)     & \textbf{0.706} & 0.956 & 0.357 \\
GP-ICM (multi-output)      & 0.607          & 0.911 & 0.286 \\
LogME \citep{YouLogME2021} & 0.527          & 0.733 & 0.143 \\
Capacity ($-\dim$)         & $-0.392$       & $-0.036$ & $-0.629$ \\
\bottomrule
\end{tabular}
\end{table}

\subsection{Residual Correlation in Frozen Image Encoders}
\label{sec:images}

To test whether the residual-correlation mechanism extends beyond tabular data, we apply the same GP framework to frozen image representations for keypoint regression. We evaluate ten pretrained encoders, including self-supervised transformers (DINOv2 ViT-S/B/L), supervised vision transformers (ViT-B/L, DeiT3-B), CNNs (ResNet-50, ConvNeXt-B), Swin-B, and CLIP. All encoder weights remain fixed, and only the GP readouts are trained.

\begin{figure}[t]
\centering
\includegraphics[width=0.58\linewidth]{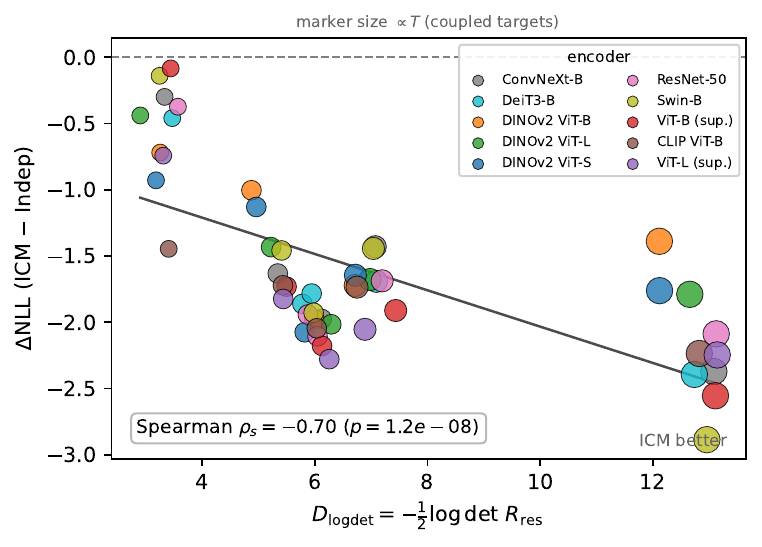}
\caption{Relationship between $D_{\text{logdet}}$ and the observed ICM benefit on frozen image representations for MPII Human Pose. Each point is an encoder--subset condition; marker size indicates the number of coupled targets. The diagnostic remains strongly associated with the uncertainty gain from coregionalisation ($\rho_s=-0.70$).}
\label{fig:mpii_dlogdet}
\end{figure}

\paragraph{MPII Human Pose.}
We construct 50 conditions by combining ten frozen encoders with keypoint subsets of varying sizes ($T=8$--$24$). Independent GPs achieve moderate predictive accuracy ($R^2_{\mathrm{Indep}}\approx0.01$--$0.27$), with substantial residual dependence remaining. Across these conditions, $D_{\text{logdet}}$ is strongly associated with the observed ICM improvement in joint NLL ($\rho_s=-0.70$, $p\approx10^{-8}$; \Cref{fig:mpii_dlogdet}). Larger target counts are also associated with greater NLL gains ($\rho_s=-0.60$, $p<10^{-5}$), consistent with the target-growth results on tabular datasets. Point-prediction changes remain small.

\paragraph{Boundary cases.}
COCO and AFLW2000-3D illustrate two limitations. On COCO, independent GPs perform poorly across encoders ($R^2_{\mathrm{Indep}}\le0.12$, median $0.02$), with little residual dependence that benefits Gaussian covariance coupling. In contrast, AFLW2000-3D has highly correlated facial landmarks and shows large joint NLL gains. However, its residual correlation matrices become nearly singular, causing $D_{\text{logdet}}$ to saturate and lose its ability to distinguish between conditions. Despite this limitation, ICM gains still increase with target count, and single-output GP evidence outperforms GP-ICM for representation selection.

Overall, the image experiments support the findings on tabular datasets. The diagnostic is most informative when independent predictors leave exploitable residual correlation, but less informative when this dependence is weak, or the residual correlation is nearly singular.

\section{Related Work}
\label{sec:related}

\paragraph{Multi-output Gaussian processes.}
Multi-output GPs model cross-target dependencies through shared covariance structures. The linear model of coregionalisation and its separable ICM special case allow information sharing across targets \citep{Bonilla2007,AlvarezLawrence2011}. Extensions include convolution-based covariance models and scalable approximations \citep{alvarez2011computationally}, heterogeneous output likelihoods \citep{moreno2018heterogeneous,gammelli2022generalized}, and input-dependent cross-output correlations \citep{wang2025nonstationary}. More general approaches to GP scalability are reviewed by \citet{LiuMOGP2018}. Rather than extending multi-output GP models, we investigate when separable coregionalisation improves point prediction or joint uncertainty.

\paragraph{Task relatedness and residual covariance.}
Sharing representations or parameters across related tasks can improve generalisation \citep{Caruana1997,EvgeniouPontil2004}, but inappropriate sharing may cause negative transfer \citep{ruder2017overview,standley2020tasks,ZhangYang2022}. The importance of modelling correlated prediction errors is well established in seemingly unrelated regression \citep{zellner1962efficient}. Building on this idea, we use residual correlation estimated from independent GPs to assess the potential joint-uncertainty benefit of coregionalisation. We further examine how this benefit differs from improvements in point prediction and representation selection.

\paragraph{Uncertainty and representation selection.}
GPs provide predictive uncertainty under Gaussian assumptions \citep{RasmussenWilliams2006}, while deep ensembles offer a practical alternative \citep{LakshminarayananEnsembles2017}. Regression calibration methods further improve the reliability of predictive uncertainty estimates \citep{kuleshov2018accurate}. Proper scoring rules, including the logarithmic score, provide a framework for evaluating predictive distributions \citep{gneiting2007strictly}. Our focus is on joint predictive uncertainty, including cross-target covariance. For representation selection, LogME uses Bayesian evidence \citep{YouLogME2021}, while H-score, LEEP, and TransRate use alternative transferability criteria \citep{BaoHScore2019,NguyenLEEP2020,huang2022transrate}. We examine whether coregionalised GP evidence improves representation rankings beyond independent GP evidence.

\section{Limitations}
\label{sec:limitations}

\paragraph{Model and comparison scope.}
Our analysis focuses on separable GP-ICM models with covariance $\mathbf{B}\otimes\mathbf{K}$ and moderate sample sizes where exact GP inference is feasible. Non-separable, sparse, inducing-point, and deep-kernel GPs may behave differently. We also fix the input representation and marginal predictor family to isolate the effect of cross-target covariance. Broader multi-output methods, including regressor chains, multi-target forests, neural joint models, and copula-based approaches, may change the marginal predictions or dependence structure and are therefore outside our controlled comparison.

\paragraph{Diagnostic scope.}
$D_{\text{logdet}}$ requires independent GP fits and residual-correlation estimation, which can be noisy or numerically unstable when residual structure is weak or highly collinear. Shrinkage improves stability in ill-conditioned settings but is not always beneficial (Appendix~\ref{app:shrinkage}). More fundamentally, the diagnostic relies on a single global Pearson correlation matrix and cannot capture input-dependent correlations that cancel when pooled or nonlinear dependence with zero Pearson correlation. These limitations are demonstrated in Appendix~\ref{app:scope}. Thus, a small $D_{\text{logdet}}$ suggests limited potential for global Gaussian covariance coupling, but does not imply statistical independence.

\paragraph{Coupling beyond covariance.}
Our conclusions about point prediction apply only to covariance-based coupling. Models that jointly learn representations or parameters across targets may improve prediction through changes to the mean function. Although a controlled neural experiment supports the residual-covariance mechanism beyond GPs (Appendix~\ref{app:neural}), it does not establish the same conclusions for representation-sharing multi-task models. Finally, the plant-phenotyping datasets contain only two targets, limiting what they can show about gains from coupling larger numbers of targets.

\section{Conclusion}
\label{sec:conclusion}
This paper examines when target coupling through GP coregionalisation improves multi-target regression. Across the separable GP-ICM settings studied, coupling provides little benefit for representation selection or point prediction but can substantially improve joint uncertainty estimation. We show that residual correlation predicts these gains much better than raw target correlation. The proposed diagnostic, $D_{\text{logdet}}$, is computed from independent GPs and is strongly associated with observed improvements in joint NLL across synthetic experiments, multi-target benchmarks, and frozen image representations.

We also introduce Residual-ICM, which incorporates residual correlation into the joint predictive covariance while preserving independent GP means and marginal variances. It achieves the best average joint NLL among the compared methods. Together, these findings show that residual dependence is more informative than raw target correlation for identifying when separable coregionalisation can improve joint uncertainty.

\acks{This work was funded by the Dutch Research Council (NWO) under grant number 482.20.700 (Twinergy) and TKI ICT (Digital Holland) under grant number PPS25-ICT26006 (ADAPT). We thank the anonymous reviewers and meta-reviewer for their constructive feedback.}

\bibliography{acml26}

\clearpage
\begin{center}
{\Large\bfseries Supplementary Material}\\[0.5em]
{\large Residual Correlation as a Diagnostic for Joint-Uncertainty Gains from GP Coregionalisation}\\[0.5em]
Fangqin Zhou and Joaquin Vanschoren
\end{center}










\appendix
\label{supplementary}



\section{Scope and Robustness of the Residual-Correlation Diagnostic}

\subsection{Scope Stress Tests}
\label{app:scope}
Because the diagnostic $D_{\text{logdet}}$ is computed from a single global Pearson correlation matrix of standardised residuals, it captures global linear residual dependence but not all forms of statistical dependence. We examine two cases outside this scope: input-dependent correlation and nonlinear dependence with zero Pearson correlation.

\paragraph{Input-dependent residual correlation.}
We generate $T=4$ targets whose residual correlation structure changes across two regions of the input space. The nonstationary setting has positive dependence in one region and negative dependence in the other, while a stationary control maintains positive dependence throughout. We evaluate Residual-ICM using either a single global correlation matrix $\mathbf{R}$ or an oracle model with a separate matrix for each region. Results are averaged over three seeds.

At $\rho=0.9$ (\Cref{tab:nonstationary}), the opposing correlations in the nonstationary setting largely cancel when pooled. Consequently, $D_{\text{logdet}}$ decreases from $1.76$ in the stationary control to $0.08$, and global Residual-ICM improves joint NLL by only $0.08$ nats. In contrast, the oracle region-wise model achieves a gain of $1.18$ nats by modelling the local dependence separately.

This result illustrates an important limitation: a small global $D_{\text{logdet}}$ does not imply that residuals are independent. It indicates that little dependence can be exploited by a single global Gaussian correlation matrix. Models with input-dependent coregionalisation \citep{wang2025nonstationary} can address dependence that varies across the input space.

\begin{table}[h]
\centering
\footnotesize
\caption{Input-dependent residual correlation ($T=4$, $\rho=0.9$, mean over three seeds). The nonstationary setting has opposing correlation patterns across two input regions, which largely cancel in the global correlation matrix. An oracle region-wise model recovers the local dependence. Negative $\dnll$ indicates improved joint NLL relative to independent predictions.}
\label{tab:nonstationary}
\setlength{\tabcolsep}{8pt}
\begin{tabular}{lccc}
\toprule
Regime & global $D_{\text{logdet}}$ & Residual-ICM (global $\mathbf{R}$) & oracle region-wise $\mathbf{R}$ \\
\midrule
stationary    & $1.76$ & $-1.86$ & $-1.86$ \\
nonstationary & $0.08$ & $-0.08$ & $\mathbf{-1.18}$ \\
\bottomrule
\end{tabular}
\end{table}

\paragraph{Nonlinear, Pearson-uncorrelated dependence.}
We test whether the diagnostic can capture nonlinear residual dependence that has zero Pearson correlation. We construct standardised residuals satisfying
$z_2=(z_1^2-1)/\sqrt{2}$,
where $z_1$ follows a standard Gaussian distribution. Although $z_1$ and $z_2$ are statistically dependent, their Pearson correlation is zero. This dependence can instead be detected using distance correlation \citep{szekely2007measuring}.

As shown in \Cref{tab:nonlinear}, the nonlinear setting has near-zero Pearson correlation ($0.01$) and $D_{\text{logdet}}\approx0$, despite a distance correlation of $0.33$. Gaussian covariance coupling provides no improvement in joint NLL, whereas a cross-validated nonparametric conditional model improves it by $0.30$ nats. In the linear-dependence control, Gaussian covariance coupling improves joint NLL by $0.23$ nats, consistent with the nonzero diagnostic.

These results clarify the scope of $D_{\text{logdet}}$. It measures the potential benefit of modelling global linear residual correlation with a Gaussian covariance structure, rather than statistical dependence in general. A small diagnostic value therefore does not rule out gains from models that capture nonlinear or input-dependent dependence.

\begin{table}[t]
\centering
\footnotesize
\caption{Nonlinear residual-dependence control ($T=2$, mean over three seeds). The nonlinear setting has near-zero Pearson correlation but nonzero distance correlation. Gaussian covariance coupling provides no improvement, whereas a nonparametric conditional model captures the remaining dependence. Negative $\dnll$ favours the coupled model.}
\label{tab:nonlinear}
\setlength{\tabcolsep}{6pt}
\begin{tabular}{lccccc}
\toprule
Residual coupling & $D_{\text{logdet}}$ & Pearson $|\rho_p|$ & Dist.\ corr. & Gaussian $\dnll$ & $k$-NN cond.\ $\dnll$ \\
\midrule
independent & $0.00$ & $0.02$ & $0.10$ & $-0.00$ & $+0.05$ \\
linear      & $0.22$ & $0.60$ & $0.56$ & $\mathbf{-0.23}$ & $-0.18$ \\
nonlinear   & $0.00$ & $0.01$ & $0.33$ & $\phantom{-}0.00$ & $\mathbf{-0.30}$ \\
\bottomrule
\end{tabular}
\end{table}

\subsection{Robustness of the Diagnostic Association}
\label{app:robustness}
The main diagnostic association is evaluated across 16 datasets. To assess its stability, we examine sensitivity to dataset sampling, individual datasets, random seeds, and the definition of residual correlation. Specifically, we use 10,000 bootstrap resamples of the datasets, leave-one-dataset-out (LODO) analysis, correlations computed separately for each of five seeds, and a one-sided permutation test.

We repeat these analyses using both predictive-standardised residual correlation ($\mathbf{R}_z$), which is the default estimator in this paper, and raw residual correlation ($\mathbf{R}_{\mathrm{raw}}$). This comparison tests whether standardising residuals by their predictive uncertainties materially affects the observed association.

As shown in \Cref{tab:robustness}, the association remains strongly negative under both residual definitions. For the predictive-standardised estimator, Spearman correlation is $-0.83$, with a bootstrap 95\% confidence interval of $[-0.99,-0.47]$. Removing any single dataset gives correlations between $-0.92$ and $-0.80$, while per-seed correlations range from $-0.89$ to $-0.73$. The one-sided permutation test gives $p<10^{-4}$. Raw residual correlation produces similar results, with Spearman $\rho_s=-0.85$, a bootstrap interval of $[-0.98,-0.51]$, and no sign reversal in the LODO or per-seed analyses.

Together, these results indicate that the association between residual correlation and ICM uncertainty gains is robust to the tested sources of variation. In particular, it is not driven by a single dataset or seed, and its strength is similar under raw and predictive-standardised residual correlation.

\begin{table}[t]
\centering
\footnotesize
\caption{Robustness of the association between $D_{\text{logdet}}$ and observed ICM gains ($\dnll$) across 16 datasets. Results compare predictive-standardised residual correlation ($\mathbf{R}_z$) with raw residual correlation ($\mathbf{R}_{\mathrm{raw}}$). More negative correlations indicate a stronger association between the diagnostic and improved joint NLL.}
\label{tab:robustness}
\setlength{\tabcolsep}{8pt}
\begin{tabular}{lcc}
\toprule
Check & $\mathbf{R}_z$ (paper) & $\mathbf{R}_{\text{raw}}$ (sensitivity) \\
\midrule
Spearman $\rho_s$                              & $-0.83$ & $-0.85$ \\
Pearson $\rho_p$                               & $-0.91$ & $-0.93$ \\
Bootstrap 95\% CI (over datasets)              & $[-0.99,\,-0.47]$ & $[-0.98,\,-0.51]$ \\
Leave-one-dataset-out range                    & $[-0.92,\,-0.80]$ & $[-0.91,\,-0.81]$ \\
Per-seed range (5 seeds)                       & $[-0.89,\,-0.73]$ & $[-0.87,\,-0.67]$ \\
Permutation $p$ (one-sided)                    & $<10^{-4}$ & $<10^{-4}$ \\
\bottomrule
\end{tabular}
\end{table}

\subsection{Stabilising $D_{\text{logdet}}$ under Residual Collinearity}
\label{app:shrinkage}
The diagnostic $D_{\text{logdet}}=-\tfrac12\log\det\mathbf{R}_{\text{res}}$ grows without bound as the residual correlation matrix approaches singularity. The empirical matrix can become singular or nearly singular for two reasons. First, when the number of residual samples is smaller than the number of targets ($n<T$), the sample correlation matrix is rank-deficient, with rank at most $n-1$. Second, even when $n>T$, strong collinearity among targets can produce very small eigenvalues. In either case, the determinant may become numerically indistinguishable from zero. When it is clipped to a numerical floor, the diagnostic saturates and can no longer distinguish different levels of residual dependence.

The AFLW face-landmark experiments illustrate the second case. Although all conditions have more samples than targets ($n=200$, $T\leq40$), the residuals are highly collinear. Consequently, 40 of the 50 conditions reach the numerical ceiling of $-\tfrac12\log(10^{-12})\approx13.8$ under the empirical estimator, even though their observed ICM gains vary substantially ($\dnll$ from $-8$ to $-38$).

To address this problem, we apply Ledoit--Wolf covariance shrinkage \citep{ledoit2004well}. The empirical residual covariance is replaced by
$\widehat{\boldsymbol{\Sigma}}_\lambda=(1-\lambda)\widehat{\boldsymbol{\Sigma}}+\lambda\,\tfrac{\operatorname{tr}\widehat{\boldsymbol{\Sigma}}}{T}\mathbf{I}$,
where the shrinkage intensity $\lambda$ is estimated analytically from the data, without a tuning parameter. We then convert the shrunk covariance to a correlation matrix and compute $D_{\text{logdet}}$. Shrinkage regularises small eigenvalues and can prevent numerical saturation. However, the resulting diagnostic is based on a regularised estimate of residual correlation, so its value no longer corresponds exactly to the idealised gain computed from the original empirical correlation matrix.

\begin{figure}[t]
\centering
\includegraphics[width=\textwidth]{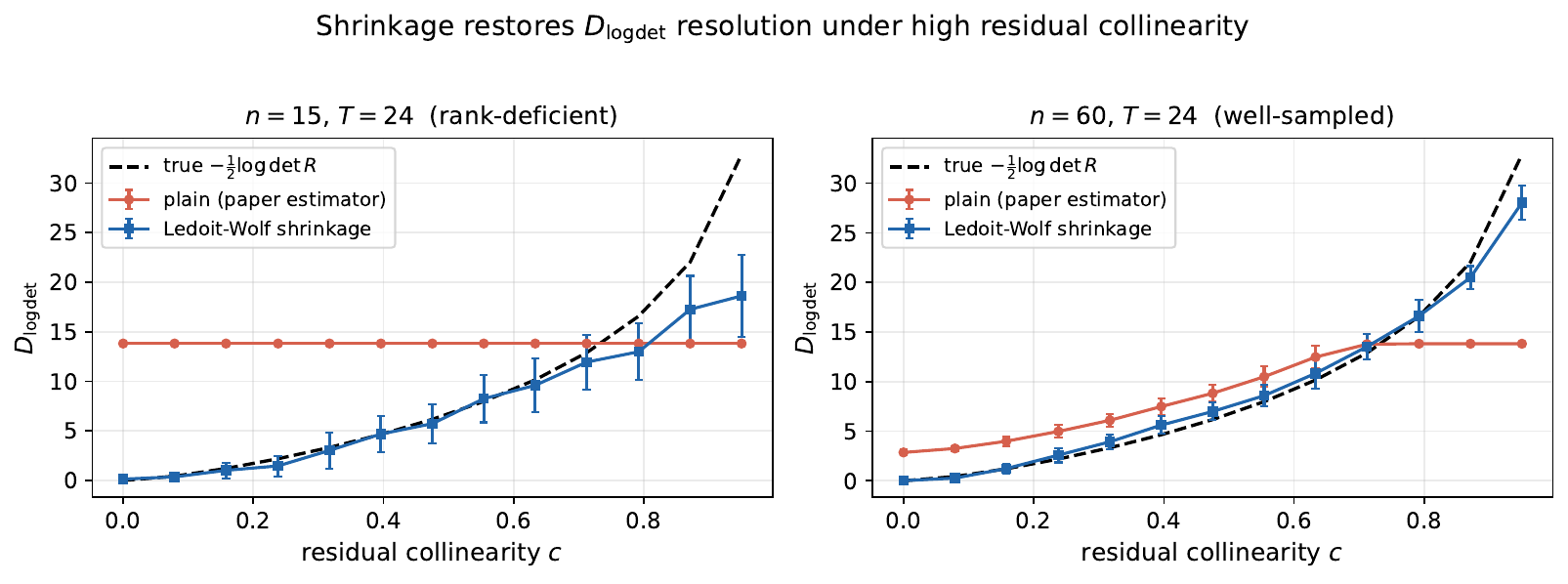}
\caption{Controlled residual-collinearity sweep ($T=24$, equicorrelation $c$, 20 seeds). When $n<T$ (left), the empirical correlation matrix is rank-deficient, causing the plain $D_{\text{logdet}}$ estimator (red) to saturate at its numerical ceiling. Ledoit--Wolf shrinkage (blue) restores variation with correlation strength. When $n>T$ (right), strong collinearity can still cause numerical saturation, which shrinkage mitigates. Dashed lines show the population value of $D_{\text{logdet}}$. The latter regime illustrates the instability observed in AFLW ($n=200$, $T\leq40$).}
\label{fig:shrinkage_synth}
\end{figure}

\paragraph{Controlled verification.}
We first test shrinkage using synthetic residuals drawn from an equicorrelation model,
$\mathbf{R}(c)=(1-c)\mathbf{I}+c\,\mathbf{1}\mathbf{1}^\top$,
where $c$ controls the correlation strength. We fix $T=24$ and vary $c$ under two sample-size regimes (\Cref{fig:shrinkage_synth}). When $n=15<T$, the empirical correlation matrix is structurally rank-deficient. Its determinant reaches the numerical floor for every value of $c$, leaving the unregularised diagnostic constant and unable to distinguish correlation levels. Ledoit--Wolf shrinkage restores variation in the diagnostic, which increases with the underlying correlation strength (Spearman $\rho_s=+0.72$ in the high-collinearity range). When $n=60>T$, the empirical correlation matrix is generally full-rank, but becomes poorly conditioned as $c$ increases. Shrinkage delays numerical saturation and extends the range over which the diagnostic can distinguish different correlation strengths. This is the same type of collinearity-driven instability observed in AFLW, although AFLW has a larger sample size.

\paragraph{Effect on the real diagnostic.}
We compare the empirical and shrunk diagnostics on the 16 tabular datasets and the MPII and AFLW keypoint conditions (\Cref{tab:shrinkage}). For a consistent comparison, we use the same seed-0 run for both estimators. The resulting correlations therefore differ slightly from the multi-seed results reported in the main text.

Shrinkage improves the association between $D_{\text{logdet}}$ and observed ICM gains on AFLW, where Spearman $\rho_s$ changes from $-0.69$ to $-0.83$. The association remains nearly unchanged on MPII ($-0.70$ to $-0.71$). However, shrinkage weakens the association across the tabular datasets ($-0.81$ to $-0.60$). The tabular degradation is largely driven by \texttt{rf1}. Shrinkage reduces its diagnostic from $1.42$ to $0.25$, despite a substantial observed ICM gain ($\dnll=-1.37$), causing it to be ranked too low relative to other datasets. This effect is specific to the predictive-standardised residual estimator. When using raw residuals, the corresponding Spearman correlation changes only from $-0.78$ to $-0.76$ after shrinkage. The degradation with standardised residuals is also observed across all five seeds.

These results show that shrinkage is useful when the empirical residual correlation matrix is ill-conditioned or numerically saturated, but may distort the diagnostic ranking when the original estimate is already stable. Therefore, we retain the empirical estimator by default and apply shrinkage only when numerical stabilisation is needed.

\begin{table}[t]
\centering
\footnotesize
\caption{Spearman correlation $\rho_s(D_{\text{logdet}},\dnll)$ under empirical and Ledoit--Wolf shrinkage estimators. Results use the same seed-0 run for the 16 tabular datasets and the MPII and AFLW keypoint conditions. Shrinkage improves the diagnostic association on AFLW, has little effect on MPII, and weakens the association across tabular datasets.}
\label{tab:shrinkage}
\setlength{\tabcolsep}{8pt}
\begin{tabular}{lccc}
\toprule
Regime & \#conditions & plain & Ledoit--Wolf \\
\midrule
Tabular benchmarks            & 16 & $\mathbf{-0.81}$ & $-0.60$ \\
MPII (moderate residual)      & 50 & $-0.70$ & $-0.71$ \\
AFLW (collinear, saturated)   & 50 & $-0.69$ & $\mathbf{-0.83}$ \\
\bottomrule
\end{tabular}
\end{table}

\section{Independent-GP Baseline Checks}

\subsection{Calibration of the Single-Output GP Baseline}
\label{app:calibration}

We first evaluate whether independent GPs provide a competitive and well-calibrated baseline. We compare them with Bayesian linear regression (BLR) and deep ensembles on three datasets covering nonlinear (Tecator and Energy) and near-linear (PHO) cross-target relationships. All targets are standardised, and results are averaged over both prediction directions for each dataset.

As shown in \Cref{tab:calibration}, independent GPs achieve the lowest NLL and narrowest predictive intervals on Tecator and Energy, with coverage gaps comparable to the other methods. On PHO, the GP and deep ensemble achieve similar NLL (0.918 and 0.913, respectively), both outperforming BLR, with comparable coverage. These results establish independent GPs as a competitive baseline for evaluating whether modelling residual dependence further improves joint uncertainty.

\begin{table}[t]
\centering
\scriptsize
\caption{Cross-target predictive uncertainty of independent GPs, Bayesian linear regression (BLR), and deep ensembles. Results are averaged over both prediction directions per dataset, with standardised targets. Lower NLL and coverage gaps are better; lower sharpness indicates narrower intervals and should be interpreted alongside coverage. Tecator and Energy have nonlinear cross-target relationships, while PHO is near-linear.}
\label{tab:calibration}
\setlength{\tabcolsep}{5pt}
\begin{tabular}{llcccc}
\toprule
Dataset & Method & NLL $\downarrow$ & Gap$_{90}$ $\downarrow$ & Gap$_{95}$ $\downarrow$ & Sharp $\downarrow$ \\
\midrule
\multirow{3}{*}{Tecator}
  & GP (RBF) & $\mathbf{-0.790}$ & 0.027 & 0.012 & \textbf{0.165} \\
  & BLR      & $-0.475$          & 0.010 & 0.006 & 0.174 \\
  & Ensemble & $\phantom{-}$0.277 & 0.073 & 0.042 & 0.434 \\
\midrule
\multirow{3}{*}{Energy}
  & GP (RBF) & \textbf{$\phantom{-}$0.262} & 0.041 & 0.023 & \textbf{0.206} \\
  & BLR      & $\phantom{-}$0.323          & 0.017 & 0.020 & 0.316 \\
  & Ensemble & $\phantom{-}$0.327          & 0.039 & 0.014 & 0.280 \\
\midrule
\multirow{3}{*}{PHO}
  & GP (RBF) & $\phantom{-}$0.918          & 0.026 & 0.026 & 0.532 \\
  & BLR      & $\phantom{-}$0.959          & 0.023 & 0.034 & 0.546 \\
  & Ensemble & \textbf{$\phantom{-}$0.913} & 0.033 & 0.036 & \textbf{0.499} \\
\bottomrule
\end{tabular}
\end{table}

\subsection{Cross-Target Recovery: Single-Output GP vs.\ Probes}
\label{app:gateb}
We next examine how well independent GPs recover cross-target relationships. Using a frozen representation learned for one target, we predict the other target with a GP or a linear probe. This comparison tests whether the GP can recover cross-target information beyond that captured by a linear predictor.

\Cref{tab:gateb} shows that the GP outperforms the linear probe on PHO and Energy, with bootstrap 95\% confidence intervals excluding zero. On the smaller SH and WUR datasets ($n\leq50$), the differences are not statistically distinguishable from zero. We also evaluate a one-hidden-layer MLP probe with 64 hidden units. It performs worse than the linear probe across all tested datasets and directions (e.g., $R^2=0.548$ vs.\ $0.587$ on PHO and $0.886$ vs.\ $0.921$ on Energy), further supporting the GP's advantage in the larger datasets.

These experiments use the full PHO ($n=661$) and Energy ($n=768$) datasets, unlike the subsampled independent-GP versus ICM comparisons in the main text. They provide a separate check of the GP's ability to recover cross-target structure, supporting its use as the independent predictor in our residual-correlation analysis.

\begin{table}[t]
\centering
\footnotesize
\caption{Cross-target recovery of independent GPs relative to linear probes, measured by $\Delta R^2=R^2_{\text{GP}}-R^2_{\text{linear probe}}$. Positive values favour the GP. Bootstrap 95\% confidence intervals are estimated from 2000 samples. Improvements are significant on PHO and Energy but not on the smaller SH and WUR datasets.}
\label{tab:gateb}
\setlength{\tabcolsep}{7pt}
\begin{tabular}{llccc}
\toprule
Dataset & Direction & $\Delta R^2$ & 95\% CI & $n$ \\
\midrule
PHO    & $\vcmax \to \jmax$          & $+0.032$ & $[+0.015,\;+0.060]$ & 661 \\
PHO    & $\jmax \to \vcmax$          & $+0.031$ & $[+0.017,\;+0.048]$ & 661 \\
Energy & cooling $\to$ heating       & $+0.074$ & $[+0.047,\;+0.106]$ & 768 \\
SH     & $\vcmax \to \jmax$          & $+0.076$ & $[-0.292,\;+0.507]$ & 50  \\
WUR    & $\vcmax \to \jmax$          & $-0.071$ & $[-0.249,\;+0.167]$ & 48  \\
\bottomrule
\end{tabular}
\end{table}

\section{Synthetic Phase Diagram: Generator and Sensitivity Analysis}
\label{app:synthetic}

We use a latent-factor generator to simulate high-dimensional observations whose underlying signal has low intrinsic dimension. For each experimental setting, we sample $r=3$ shared latent factors $\mathbf{z}\in\mathbb{R}^{n\times r}$ from a standard Gaussian distribution and heterogeneous target loadings $\mathbf{L}\in\mathbb{R}^{T\times r}$. The observed features are generated through a nonlinear transformation,
$\mathbf{X}=\tanh(\mathbf{z}\mathbf{W}_{\text{in}})$
where $\mathbf{W}_{\text{in}}\in\mathbb{R}^{r\times p}$. This construction allows the ambient feature dimension $p$ to vary while keeping the intrinsic signal dimension fixed at $r=3$. The targets are generated as
$\mathbf{Y}=\operatorname{std}(\mathbf{z}\mathbf{L}^\top)+\sigma_\epsilon\,\boldsymbol{\epsilon}$ 
where $\sigma_\epsilon=0.6$ and $\boldsymbol{\epsilon}\sim\mathcal{N}(0,\mathbf{I})$.

We fix $n=150$ and vary the number of targets $T\in\{2,3,5,8\}$ and feature dimensions $p\in\{40,150,600\}$, corresponding to $\pn\in\{0.27,1.0,4.0\}$. Each setting is evaluated over five random seeds using the same independent-GP and ICM configurations as in the main text. This design examines how target count and ambient feature dimensionality affect the benefit of coupling, while keeping the underlying signal dimension fixed. Cross-target correlation arises from the shared latent factors and target loadings rather than from a prescribed correlation matrix.

\begin{figure}[htbp]
\centering
\includegraphics[width=\textwidth]{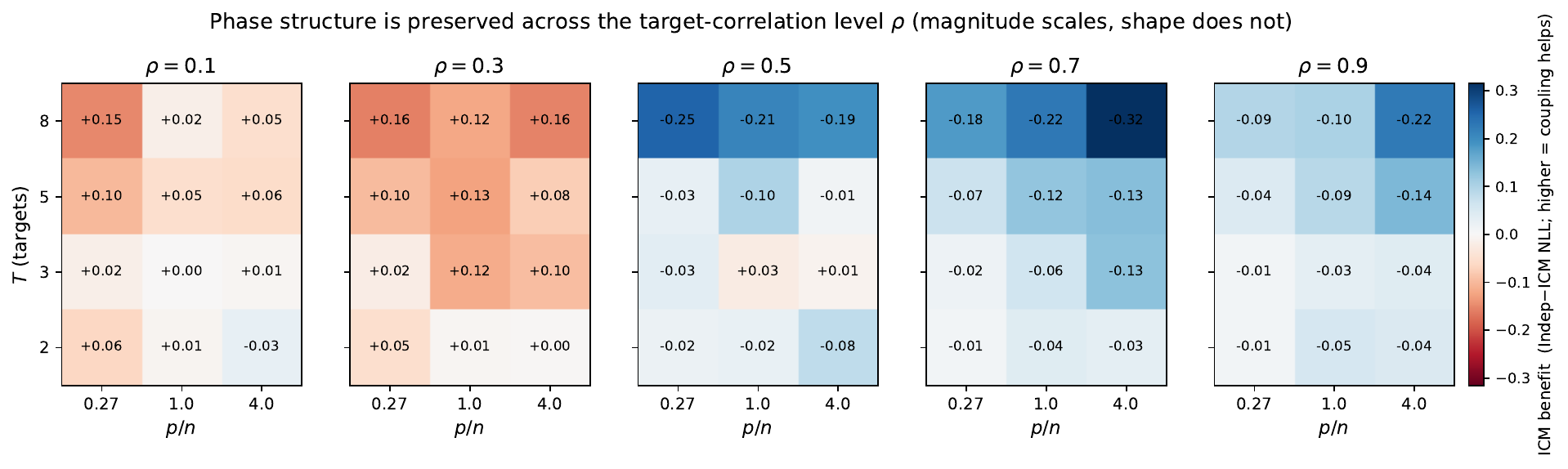}
\caption{Synthetic $(T,\pn)$ phase diagrams at five levels of shared target structure $\rho$. Cell values show $\dnll=\mathrm{NLL}_{\mathrm{ICM}}-\mathrm{NLL}_{\mathrm{Indep}}$, where negative values favour ICM. Colours represent $-\dnll$, with blue indicating larger coupling benefits. At $\rho\leq0.3$, ICM provides little benefit. At higher $\rho$, the benefit becomes concentrated in settings with larger $T$ and $\pn$.}
\label{fig:phase_rho}
\end{figure}

\paragraph{Robustness to target-correlation strength.}
We test whether the phase diagram depends on the level of shared target structure. To control this, we introduce a parameter $\rho$ into the target generator:
$\mathbf{Y}_c=\sqrt{\rho}\,\operatorname{std}(\mathbf{z}\mathbf{L}^\top)+\sqrt{1-\rho}\,\mathbf{o}$
where $\mathbf{o}\sim\mathcal{N}(0,\mathbf{I})$ contains independent noise for each target. Consequently, increasing $\rho$ increases the contribution of the shared latent component and the magnitude of the resulting target correlations. We repeat the $(T,\pn)$ sweep for $\rho\in\{0.1,0.3,0.5,0.7,0.9\}$ (\Cref{fig:phase_rho}).

\Cref{fig:phase_rho} shows two main patterns. First, when shared structure is sufficiently strong ($\rho\geq0.5$), ICM gains are generally larger at higher $T$ and $\pn$, consistent with the main phase diagram. The mean $\dnll$ across the grid is $+0.04$, $+0.09$, $-0.08$, $-0.11$, and $-0.07$ for increasing values of $\rho$. Thus, coupling becomes beneficial as shared structure increases from weak to moderate levels, although the average gain is not strictly monotonic, particularly at $\rho=0.9$.

Second, when shared structure is weak ($\rho\leq0.3$), ICM provides no average improvement and slightly worsens joint NLL. This suggests that coupling offers little benefit when independent predictors leave insufficient correlated residual structure. At $\rho=0.7$ and $0.9$, the residual diagnostic remains associated with the observed ICM gains, with Spearman correlations of $-0.58$ and $-0.53$, respectively. In contrast, raw target correlation varies little within each fixed-$\rho$ setting and provides limited information for ranking the experimental conditions.

Overall, these results suggest that the phase structure is not specific to one level of target correlation. The benefit of coupling depends on the residual dependence remaining after independent prediction, rather than on the strength of target correlation alone.

\section{Covariance-Modelling Ablations and Comparisons}

\subsection{ICM Configuration and Optimisation Details}
\label{app:hardening}

In the main experiments, we fix the shared ICM lengthscale $\ell$ to the median optimised lengthscale of the independent GPs and optimise only the coregionalisation matrix $\mathbf{B}$ and observation noise $\sigma^2$. We constrain $\sigma^2\in[10^{-4},0.5]$. Since the targets are standardised to unit variance, the upper bound limits how much variation the model can assign to observation noise, while the lower bound prevents near-zero noise estimates. The upper bound is active only on \texttt{wq}, where removing it improves ICM from $\dnll=-0.14$ to $-0.34$. All other datasets change by less than $0.03$.

For separable ICM, Kronecker eigendecomposition reduces the cost of computing the marginal likelihood from $O((nT)^3)$ to $O(n^3+T^3)$ by diagonalising the input and output covariance matrices separately. Our implementation uses NumPy/SciPy eigendecompositions and L-BFGS-B to optimise the coregionalisation and noise parameters.

\paragraph{Effect of stronger ICM configurations.}
The default ICM configuration uses a fixed lengthscale, bounded observation noise, and rank-1 coregionalisation. These restrictions allow a controlled comparison with independent GPs but may limit ICM performance. To assess their effects, we relax each restriction separately (\Cref{tab:hardening}).

All three changes improve mean ICM performance. Re-optimising the shared lengthscale improves 15 of 16 datasets, while increasing the rank reduces mean $\dnll$ from $-0.62$ at rank 1 to $-0.82$ at rank 2 and $-0.93$ at rank $T{-}1$. Removing the noise cap has little average effect, as the upper bound is active only on \texttt{wq}. These results show that the default configuration is relatively restrictive and does not favour ICM in the comparison.

However, stronger ICM configurations do not resolve all failures. Increasing the rank improves \texttt{sarcos} from $\dnll=+0.44$ to $-0.06$, but \texttt{tecator} still degrades joint NLL ($+0.78\to+0.62$), as does \texttt{edm} ($+0.64$ at every tested rank). These datasets illustrate different limitations. Tecator is a high-accuracy case in which independent GPs leave limited residual structure to exploit. Sarcos retains substantial residual dependence ($D_{\text{logdet}}\approx1.0$), but rank-1 ICM worsens joint NLL, whereas Residual-ICM improves it ($\dnll=-0.83$). This suggests that preserving the independent GP means and marginal variances can be beneficial when jointly fitting the ICM degrades performance. EDM, in contrast, has little exploitable residual dependence.

\begin{table}[h]
\centering
\footnotesize
\caption{ICM configuration ablation across 16 datasets and five seeds. Negative $\dnll$ favours ICM. Removing the noise cap, increasing the rank, or re-optimising the shared lengthscale improves mean ICM performance. The default rank-1 mean ($-0.616$) differs slightly from \Cref{tab:method} ($-0.614$) because the ICM fit for SH (seed 2) converged to a different optimum in this independent re-run.}
\label{tab:hardening}
\setlength{\tabcolsep}{7pt}
\begin{tabular}{p{4.8cm}ccccc}
\toprule
Variant
& Default $r=1$
& No cap
& $r=2$
& $r=T{-}1$
& Optimised $\ell$ \\
\midrule
Mean $\dnll$
& $-0.616$
& $-0.627$
& $-0.822$
& $\mathbf{-0.928}$
& $-0.814$ \\
\# datasets improving over default
& ---
& 3/16
& 14/16
& 8/16
& 15/16 \\
\bottomrule
\end{tabular}
\end{table}

\subsection{Does a Stronger ICM Close the Gap to Residual-ICM?}
\label{app:rankcomp}
We further compare ICM across different ranks with Residual-ICM (\Cref{tab:rank_resicm}). As an upper-bound reference, we include a hindsight oracle that selects the best tested ICM rank for each dataset using its evaluation $\dnll$. This oracle is not a deployable selection strategy because it uses held-out evaluation results. We report results across all 16 datasets and separately for the ten datasets with $T\geq3$, since rank $T{-}1$ coincides with rank 1 when $T=2$.

Increasing the ICM rank improves mean $\dnll$ from $-0.62$ at rank 1 to $-0.93$ at rank $T{-}1$. Hindsight rank selection gives a slightly larger improvement ($-0.94$). Nevertheless, Residual-ICM achieves better average joint NLL, with $\dnll=-1.13$ across all datasets and $-1.54$ for $T\geq3$.

Residual-ICM also shows fewer and smaller degradations relative to independent GPs. Summing positive $\dnll$ over datasets where coupling worsens performance, hindsight-rank ICM accumulates $1.26$ nats across all datasets and $0.62$ nats for $T\geq3$, compared with $0.10$ and $0.01$ for Residual-ICM. Although increasing the ICM rank resolves the failure on \texttt{sarcos}, the degradations on \texttt{tecator} and \texttt{edm} remain.

These results show that stronger ICM configurations improve joint uncertainty estimates but do not fully close the gap to Residual-ICM. Because Residual-ICM does not depend on the ICM rank, its results are identical to those reported in the main text (\Cref{tab:method}); only the ICM configurations vary in this comparison.

\begin{table}[t]
\centering
\footnotesize
\caption{ICM rank sweep compared with Residual-ICM over five seeds. We report mean $\dnll$ and the sum of positive $\dnll$ across datasets where coupling worsens joint NLL. ``Hindsight rank'' selects the best tested ICM rank per dataset using evaluation $\dnll$ and is therefore an oracle rather than a deployable selection rule.}
\label{tab:rank_resicm}
\setlength{\tabcolsep}{6pt}
\begin{tabular}{lcccc}
\toprule
 & \multicolumn{2}{c}{All 16 datasets}
 & \multicolumn{2}{c}{$T\geq3$ (10 datasets)} \\
\cmidrule(lr){2-3}\cmidrule(lr){4-5}
Method
& mean $\dnll$
& $\sum$ positive $\dnll$
& mean $\dnll$
& $\sum$ positive $\dnll$ \\
\midrule
ICM, rank 1
& $-0.62$ & $1.85$
& $-0.89$ & $1.22$ \\
ICM, rank 2
& $-0.82$ & $1.43$
& $-1.20$ & $0.79$ \\
ICM, rank $T{-}1$
& $-0.93$ & $1.26$
& $-1.38$ & $0.62$ \\
ICM, hindsight rank
& $-0.94$ & $1.26$
& $-1.40$ & $0.62$ \\
Residual-ICM
& $\mathbf{-1.13}$ & $\mathbf{0.10}$
& $\mathbf{-1.54}$ & $\mathbf{0.01}$ \\
\bottomrule
\end{tabular}
\end{table}

\subsection{Sensitivity of the Diagnostic Gate to the Accuracy Ceiling}
\label{app:ceiling}
The diagnostic gate in \S\ref{sec:method} fits ICM only when $D_{\text{logdet}}>\tau$ and the independent GP has predictive accuracy below a ceiling, $R^2_{\mathrm{Indep}}<c$. The main experiments use $c=0.9$ to avoid coupling targets that are already predicted accurately. Here, we examine the sensitivity of the gate to this threshold. We vary $c$ while fixing $\tau=0.5$, which gives a mean $\dnll=-0.64$ at $c=0.9$, close to the main-text result of $-0.61$ (\Cref{tab:method}). Unlike the main experiment, where $\tau$ is selected by leave-one-dataset-out (LODO) cross-validation, this analysis uses a fixed value to isolate the effect of the accuracy ceiling. The two approaches make identical gate decisions on 13 of the 16 datasets.

As shown in \Cref{tab:ceiling}, the gate is relatively insensitive to the accuracy ceiling. Mean $\dnll$ remains at $-0.641$ for $c\in[0.85,0.926]$, with the default $c=0.9$ lying within this range. A lower ceiling makes the gate more restrictive, excluding datasets that benefit from ICM, such as \texttt{rf2} and \texttt{scm1d}. At $c=0.80$, mean $\dnll$ worsens to $-0.588$.

Increasing the ceiling to $c=0.93$ allows the gate to include \texttt{rf1} ($R^2_{\mathrm{Indep}}=0.927$, $\dnll=-1.09$), giving the lowest mean $\dnll$ in this sweep ($-0.709$). However, raising it further also includes \texttt{sarcos} ($R^2_{\mathrm{Indep}}=0.933$), where ICM worsens joint NLL ($\dnll=+0.44$). Consequently, mean $\dnll$ increases to $-0.682$. The apparent optimum at $c=0.93$ is identified from the evaluation results and should not be interpreted as a generally optimal threshold.

We also examine whether the gate depends on how residual correlation is estimated. Using raw instead of predictive-standardised residuals produces similar associations with observed ICM gains: Spearman $\rho_s=-0.85$ versus $-0.83$, per-target-normalised correlations of $-0.89$ versus $-0.85$, and partial correlations of $-0.65$ versus $-0.61$ after controlling for $T$, $\pn$, and $R^2_{\mathrm{Indep}}$. All gate decisions remain unchanged. We retain predictive-standardised residuals because they are consistent with the covariance construction used in Residual-ICM,
$\boldsymbol{\Sigma}(x)=\operatorname{diag}(\mathbf{s})\mathbf{R}_{\text{res}}\operatorname{diag}(\mathbf{s})$.

Overall, the accuracy ceiling serves as a practical screening threshold rather than a theoretically determined constant. The default $c=0.9$ is a conservative choice within a stable range. Across all tested ceilings, the gate achieves mean $\dnll$ between $-0.59$ and $-0.71$, compared with $-1.13$ for Residual-ICM. Thus, the main conclusion is insensitive to the exact ceiling: Residual-ICM remains the stronger choice for joint uncertainty estimation, while the gate provides a simple rule for deciding when to fit a jointly coregionalised GP.

\begin{table}[t]
\centering
\footnotesize
\caption{Sensitivity of the diagnostic gate to the accuracy ceiling $c$, with $\tau=0.5$ fixed. Mean $\dnll$ is evaluated across all 16 datasets; lower values are better. Performance is unchanged for the tested ceilings from $0.85$ to $0.926$, including the main-text choice $c=0.9$. Residual-ICM achieves a lower mean $\dnll$ ($-1.13$) than the gate at every tested ceiling.}
\label{tab:ceiling}
\setlength{\tabcolsep}{6pt}
\begin{tabular}{lccccccc}
\toprule
Ceiling $c$ & $0.80$ & $0.85$ & $0.90$ & $0.926$ & $0.93$ & $0.935$ & $1.0$ \\
\midrule
Gate mean $\dnll$ & $-0.588$ & $-0.641$ & $\underline{-0.641}$ & $-0.641$ & $\mathbf{-0.709}$ & $-0.682$ & $-0.682$ \\
\bottomrule
\end{tabular}
\end{table}

\subsection{Residual Covariance in Neural Predictors}
\label{app:neural}
To test whether the residual-correlation mechanism extends beyond GPs, we evaluate covariance modelling in a neural predictor on the same 16 datasets, using three seeds. We compare three variants based on the same encoder: a diagonal predictive covariance baseline, a jointly learned full covariance, and a post-hoc residual covariance estimated from the diagonal ensemble's out-of-fold predictive-standardised residuals. We also recompute $D_{\text{logdet}}$ from the neural residuals to test whether it predicts the observed uncertainty gains.

As shown in \Cref{tab:neural}, post-hoc residual covariance improves joint NLL on 13 of 16 datasets, with a mean $\dnll=-0.80$ across all 16. The gains are strongly associated with the neural residual-correlation diagnostic ($\rho_s=-0.88$, $p<10^{-4}$). Because this approach modifies only the predictive covariance, point predictions remain unchanged. Jointly learning the full covariance produces a smaller average improvement ($\dnll=-0.34$), with no significant association between the diagnostic and NLL gains ($\rho_s=-0.25$, $p=0.36$). It also reduces $R^2$ on 12 of 16 datasets, with a mean $\dr=-0.066$.

These findings suggest that the benefit of modelling residual covariance is not specific to GP inference. However, the diagnostic is less informative when covariance is learned jointly with the mean, as changes in predictive accuracy can also affect joint NLL. This extension is limited to one neural architecture and does not establish the same behaviour for neural predictors more generally.

\begin{table}[t]
\centering
\footnotesize
\caption{Neural covariance comparison across 16 datasets and three seeds. $\dnll$ is measured relative to the diagonal deep ensemble; negative values favour covariance modelling. The diagnostic is computed from the ensemble's out-of-fold predictive-standardised residuals.}
\label{tab:neural}
\setlength{\tabcolsep}{6pt}
\begin{tabular}{lcccc}
\toprule
Variant & mean $\dnll$ & improved & $\rho_s$ with neural $D_{\text{logdet}}$ & mean $\dr$ \\
\midrule
post-hoc residual    & $\mathbf{-0.80}$ & 13/16 & $\mathbf{-0.88}$ ($p<10^{-4}$) & $0$ by construction \\
jointly learned full & $-0.34$ & 11/16 & $-0.25$ ($p=0.36$) & $-0.066$ (12/16 reduced) \\
\bottomrule
\end{tabular}
\end{table}





\section{Representation Selection: Evidence Decomposition and Results}
\label{app:proof}

We state the selection claim precisely. Let $\mathcal{L}(\mathbf{z}; \mathbf{K}, \lambda, \sigma^2) = -\tfrac12[\mathbf{z}^\top(\lambda\mathbf{K}+\sigma^2\mathbf{I})^{-1}\mathbf{z} + \log\det(\lambda\mathbf{K}+\sigma^2\mathbf{I}) + n\log 2\pi]$ be the single-output GP log-evidence for targets $\mathbf{z}$ under input kernel $\mathbf{K}$ and signal variance $\lambda$.

\begin{proposition}[Evidence decomposition for separable GP-ICM]
\label{prop:selection}
Consider the separable coregionalisation model $\operatorname{vec}(\mathbf{Y}) \sim \mathcal{N}(\mathbf{0},\ \mathbf{B}\otimes\mathbf{K}_\Phi + \sigma^2\mathbf{I}_{nT})$ with a shared input kernel $\mathbf{K}_\Phi$ and homoscedastic noise $\sigma^2$, and let $\mathbf{B}=\mathbf{U}\mathbf{\Lambda}\mathbf{U}^\top$ be the eigendecomposition of the coregionalisation matrix. Then:
\begin{enumerate}
\item[(i)] the multi-output (MO) log-evidence decomposes exactly into single-output evidences on the eigen-targets $\mathbf{Y}\mathbf{u}_t$,
\begin{equation}
\log p_{\mathrm{MO}}(\mathbf{Y}\mid \mathbf{K}_\Phi, \mathbf{B}, \sigma^2)\;=\;\sum_{t=1}^{T}\mathcal{L}\!\left(\mathbf{Y}\mathbf{u}_t\,;\,\mathbf{K}_\Phi,\ \lambda_t,\ \sigma^2\right);
\label{eq:evidence_decomp}
\end{equation}
\item[(ii)] for a fixed coregionalisation matrix $\mathbf{B}$ shared across candidate encoders, the eigenpairs $(\mathbf{u}_t,\lambda_t)$ depend only on $\mathbf{B}$ and are therefore independent of the candidate encoder $\Phi$;
\item[(iii)] if $\mathbf{B}$ has equal eigenvalues ($\mathbf{\Lambda}=\lambda\mathbf{I}$, equivalently $\mathbf{B}=\lambda\mathbf{I}$), then $\log p_{\mathrm{MO}}(\mathbf{Y}\mid\mathbf{K}_\Phi)$ coincides with the independent-GP evidence with common signal variance $\lambda$ for every encoder, and the two therefore induce identical representation rankings.
\end{enumerate}
\end{proposition}

\paragraph{Interpretation.}
Proposition~\ref{prop:selection} shows that the evidence of a separable GP-ICM can be decomposed into single-output GP evidence over rotated targets. When the coregionalisation matrix $\mathbf{B}$ is fixed across candidate representations, the target rotation and the corresponding eigenvalues are also fixed. However, this does not guarantee identical representation rankings. Unequal eigenvalues assign different kernel scales to the rotated target directions, which can change the evidence rankings relative to independent GPs. Exact agreement is guaranteed when $\mathbf{B}=\lambda\mathbf{I}$, in which case the model reduces to independent GPs with a common signal variance.

In our experiments, $\mathbf{B}$ and the noise variance are re-optimised for each candidate representation, so the rotation and eigenvalues can also vary across representations. The empirical finding that GP-ICM does not improve over single-output GP evidence (\Cref{tab:selection_full}) is consistent with, but not implied by, the analysis. The decomposition applies to separable models. LogME is the degenerate linear-kernel case ($\mathbf{K}_\Phi=\Phi\Phi^\top$) with independent targets, where the evidence is a sum of per-target terms with target-specific prior and noise precisions. The decomposition does not generally extend to non-separable multi-output GPs.

We now prove Proposition~\ref{prop:selection}. We use the column-stacking convention $\operatorname{vec}(\mathbf{Y})\in\mathbb{R}^{nT}$ (the $T$ columns of $\mathbf{Y}\in\mathbb{R}^{n\times T}$ stacked), under which the separable covariance is $\boldsymbol{\Sigma}=\mathbf{B}\otimes\mathbf{K}_\Phi+\sigma^2\mathbf{I}_{nT}$, with $(\mathbf{B}\otimes\mathbf{K}_\Phi)_{(t,i),(t',i')}=B_{tt'}(\mathbf{K}_\Phi)_{ii'}$.

\paragraph{(i) Exact decomposition.}
Eigendecompose $\mathbf{B}=\mathbf{U}\mathbf{\Lambda}\mathbf{U}^\top$ ($\mathbf{U}$ orthogonal, $\mathbf{\Lambda}=\operatorname{diag}(\lambda_1,\dots,\lambda_T)$) and $\mathbf{K}_\Phi=\mathbf{V}\mathbf{\Gamma}\mathbf{V}^\top$ ($\mathbf{\Gamma}=\operatorname{diag}(\gamma_1,\dots,\gamma_n)$). Then
\begin{equation}
\boldsymbol{\Sigma} = (\mathbf{U}\otimes\mathbf{V})\,(\mathbf{\Lambda}\otimes\mathbf{\Gamma}+\sigma^2\mathbf{I})\,(\mathbf{U}\otimes\mathbf{V})^\top,
\end{equation}
so $\boldsymbol{\Sigma}$ has eigenvalues $\lambda_t\gamma_i+\sigma^2$ and $\log\det\boldsymbol{\Sigma}=\sum_{t,i}\log(\lambda_t\gamma_i+\sigma^2)$. For the quadratic form, set $\mathbf{M}=\mathbf{V}^\top\mathbf{Y}\mathbf{U}\in\mathbb{R}^{n\times T}$; the identity $\operatorname{vec}(\mathbf{V}^\top\mathbf{Y}\mathbf{U})=(\mathbf{U}\otimes\mathbf{V})^\top\operatorname{vec}(\mathbf{Y})$ gives
\begin{equation}
\operatorname{vec}(\mathbf{Y})^\top\boldsymbol{\Sigma}^{-1}\operatorname{vec}(\mathbf{Y})
=\sum_{t,i}\frac{M_{it}^2}{\lambda_t\gamma_i+\sigma^2}.
\end{equation}
Now consider the single-output evidence at the $t$-th eigen-target $\mathbf{z}_t=\mathbf{Y}\mathbf{u}_t$. Since $\mathbf{V}^\top\mathbf{z}_t=\mathbf{V}^\top\mathbf{Y}\mathbf{u}_t=\mathbf{M}_{\cdot t}$ (the $t$-th column of $\mathbf{M}$),
\begin{equation}
\mathcal{L}(\mathbf{z}_t;\mathbf{K}_\Phi,\lambda_t,\sigma^2)
=-\tfrac12\Big[\sum_i\frac{M_{it}^2}{\lambda_t\gamma_i+\sigma^2}+\sum_i\log(\lambda_t\gamma_i+\sigma^2)+n\log 2\pi\Big].
\end{equation}
Summing over $t=1,\dots,T$ reproduces $-\tfrac12[\,\operatorname{vec}(\mathbf{Y})^\top\boldsymbol{\Sigma}^{-1}\operatorname{vec}(\mathbf{Y})+\log\det\boldsymbol{\Sigma}+nT\log 2\pi\,]=\log p_{\mathrm{MO}}(\mathbf{Y}\mid\mathbf{K}_\Phi,\mathbf{B},\sigma^2)$, which is \eqref{eq:evidence_decomp}.

\paragraph{(ii) Encoder-independence.}
When $\mathbf{B}$ is held fixed across the candidate encoders, $\mathbf{U}$ and $\mathbf{\Lambda}$ are functions of $\mathbf{B}$ alone and do not depend on $\Phi$. Hence for any two candidates $\Phi_a,\Phi_b$ the eigen-targets $\{\mathbf{Y}\mathbf{u}_t\}$ and weights $\{\lambda_t\}$ in \eqref{eq:evidence_decomp} are the same, and only $\mathbf{K}_\Phi$ changes---exactly as in the independent score $\sum_t \mathcal{L}(\mathbf{y}_t;\mathbf{K}_\Phi,b_t,\sigma^2)$, where $b_t$ are the per-target signal variances. Thus, the target rotation and weighting remain fixed
across candidate representations. The fixed weights $\lambda_t$ can nevertheless change the ranking relative to the independent score when they are unequal, because the per-direction evidences $\mathcal{L}(\mathbf{Y}\mathbf{u}_t;\mathbf{K}_\Phi,\lambda_t,\sigma^2)$ then respond differently to a change of $\mathbf{K}_\Phi$.

\paragraph{(iii) Equal eigenvalues.}
Let $\lambda_t\equiv\lambda$ and write $\mathbf{P}=(\lambda\mathbf{K}_\Phi+\sigma^2\mathbf{I})^{-1}$. The log-determinant term of $\sum_t\mathcal{L}(\mathbf{Y}\mathbf{u}_t;\mathbf{K}_\Phi,\lambda,\sigma^2)$ is $T\log\det(\lambda\mathbf{K}_\Phi+\sigma^2\mathbf{I})$, independent of $\mathbf{U}$. The quadratic term is
\begin{equation}
\sum_{t}(\mathbf{Y}\mathbf{u}_t)^\top\mathbf{P}(\mathbf{Y}\mathbf{u}_t)
=\operatorname{tr}\!\big(\mathbf{U}^\top\mathbf{Y}^\top\mathbf{P}\mathbf{Y}\mathbf{U}\big)
=\operatorname{tr}\!\big(\mathbf{Y}^\top\mathbf{P}\mathbf{Y}\,\mathbf{U}\mathbf{U}^\top\big)
=\operatorname{tr}\!\big(\mathbf{Y}^\top\mathbf{P}\mathbf{Y}\big)
=\sum_{t}\mathbf{y}_t^\top\mathbf{P}\mathbf{y}_t,
\end{equation}
using $\mathbf{U}\mathbf{U}^\top=\mathbf{I}$. Hence $\sum_t\mathcal{L}(\mathbf{Y}\mathbf{u}_t;\mathbf{K}_\Phi,\lambda,\sigma^2)=\sum_t\mathcal{L}(\mathbf{y}_t;\mathbf{K}_\Phi,\lambda,\sigma^2)$ for every $\mathbf{K}_\Phi$: the multi-output and independent (variance-$\lambda$) evidences coincide for all encoders, so they induce identical rankings. \hfill$\blacksquare$


\paragraph{Per-task selection results.}
\Cref{tab:selection_full} reports the per-task selection agreement (Kendall $\tau$) underlying the proposition above: single-output GP-LML evidence wins or ties the multi-output GP-ICM evidence in all six dataset--direction pairs, and GP-ICM never exceeds GP-LML.

\begin{table}[htbp]
\centering
\footnotesize
\caption{Per-task Kendall $\tau$ for representation selection. GP-LML achieves the highest agreement among the three methods in five of six dataset--direction pairs. GP-ICM never outperforms GP-LML.}
\label{tab:selection_full}
\setlength{\tabcolsep}{8pt}
\small
\begin{tabular}{llccc}
\toprule
Dataset & Direction & GP-LML & GP-ICM & LogME \\
\midrule
PHO     & $\vcmax \to \jmax$         & \textbf{0.867} & 0.800 & 0.733 \\
PHO     & $\jmax \to \vcmax$         & \textbf{0.956} & 0.911 & 0.600 \\
Tecator & fat $\to$ protein          & 0.467 & 0.444 & \textbf{0.556} \\
Tecator & protein $\to$ fat          & \textbf{0.733} & 0.689 & 0.556 \\
Energy  & heating $\to$ cooling      & \textbf{0.357} & 0.286 & 0.143 \\
Energy  & cooling $\to$ heating      & \textbf{0.857} & 0.514 & 0.571 \\
\midrule
Mean    &                            & \textbf{0.706} & 0.607 & 0.527 \\
\bottomrule
\end{tabular}
\end{table}



\end{document}